\documentclass[10pt,twocolumn,letterpaper]{article}

\usepackage[pagenumbers]{cvpr} % To force page numbers, e.g. for an arXiv version

\usepackage{graphicx}
\usepackage{amsmath}
\usepackage{amssymb}
\usepackage{booktabs}
\usepackage{arydshln}
\usepackage[accsupp]{axessibility} 
\usepackage[export]{adjustbox}
\usepackage{multirow}
\usepackage[toc,page]{appendix}

\DeclareMathOperator*{\argminB}{argmin}   % Jan Hlavacek
\DeclareMathOperator{\E}{\mathbb{E}}
\newcommand{\smalljump}{\vspace{2mm}}
\newcommand{\rotText}[1]{\begin{tabular}{@{}c@{}}#1\end{tabular}}

\usepackage[dvipsnames]{xcolor}

\definecolor{cvprblue}{rgb}{0.21,0.49,0.74}
\usepackage[pagebackref,breaklinks,colorlinks,citecolor=cvprblue]{hyperref}

\definecolor{DarkGreen}{rgb}{0.0, 0.6, 0.0}

\definecolor{DarkBlue}{rgb}{0.0, 0.0, 0.8}

\definecolor{DarkOrange}{rgb}{0.9, 0.5, 0.0}

\definecolor{DarkViolet}{rgb}{0.5, 0.2, 0.5}

\definecolor{TODOColor}{rgb}{0.21,0.74,0.49}

\newcommand{\ours}{\mbox{MicKey}\xspace}

\def\paperID{2801} % *** Enter the Paper ID here
\def\confName{CVPR}
\def\confYear{2024}

\title{Matching 2D Images in 3D: Metric Relative Pose from Metric Correspondences}

\author{
Axel Barroso-Laguna$^{1}$\hspace{18pt}
Sowmya Munukutla$^{1}$\hspace{18pt}
Victor Adrian Prisacariu$^{1, 2}$\hspace{18pt}
Eric Brachmann$^{1}$\\
{\normalsize $^1$Niantic \hspace{30pt} $^2$University of Oxford}\\
{\normalsize \href{https://nianticlabs.github.io/mickey/}{https://nianticlabs.github.io/mickey/}}
}

\begin{document}
\maketitle
\begin{abstract}

% Augmented Reality (AR) is becoming more accessible as advances in technology allow for AR applications to be run almost anywhere with minimal preparation, \eg, one anchor image and one localization image. However, this also presents a challenge for algorithms to work in diverse and unconstrained environments where obtaining ground truth data is difficult. In this paper, we present a novel approach that learns to estimate metric keypoint correspondences with only metric pose supervision, by combining principles of Reinforcement Learning and a differentiable pose optimizer. We prove that no depth maps or any other additional information, \eg, SfM models or overlapping scores, are needed to train state-of-the-art keypoint detectors/descriptors. We apply our method to the Map-free task and demonstrate that our metric correspondences surpass the usability of current state-of-the-art methods. Code will be made publicly available. 

Given two images, we can estimate the relative camera pose between them by establishing image-to-image correspondences. Usually, correspondences are 2D-to-2D and the pose we estimate is defined only up to scale. Some applications, aiming at instant augmented reality anywhere, require scale-metric pose estimates
, and hence, they rely on external depth estimators to recover the scale.
We present \ours, a keypoint matching pipeline that is able to predict metric correspondences in 3D camera space. By learning to match 3D coordinates across images, we are able to infer the metric relative pose without depth measurements. Depth measurements are also not required for training, nor are scene reconstructions or image overlap information. \ours is supervised only by pairs of images and their relative poses. \ours achieves state-of-the-art performance on the Map-Free Relocalisation benchmark while requiring less supervision than competing approaches.
% The code will be made publicly available.

\end{abstract}    
\section{Introduction}
\label{sec:intro}
\begin{figure}[t]
  \centering
   \includegraphics[width=1.0\linewidth]{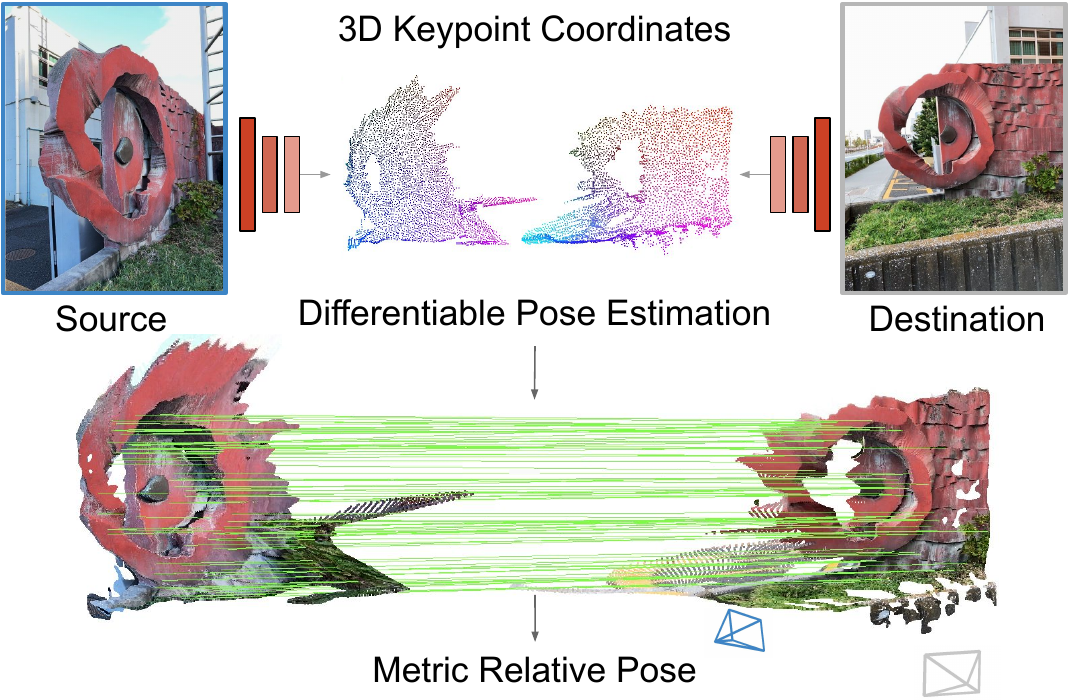}
   \caption{  
      We introduce \ours, a neural network that predicts 3D metric keypoint coordinates in camera space from a 2D input image.      
      Given two images, \ours establishes 3D-3D correspondences via descriptor matching and then applies a Kabsch \cite{kabsch1976solution} solver to recover the metric relative pose. We visualize the 3D keypoint coordinates by mapping them to the RGB cube.
   }
   \label{fig:teaser}
\end{figure}

% We introduce MicKey, a neural network that predicts 3D metric keypoint coordinates, together with their keypoint confidences and descriptors. \ours uses the camera intrinsics to lift the 3D keypoints positions to camera coordinates, where it applies a probabilistic pose solver that returns the  relative metric transformations between the two input images.  

Whether you prefer inches or centimeters, we measure and understand the world in scale-metric units. Unfortunately, the scale-metric quality of the world is lost when we project it to the image plane. The scale-ambiguity is one aspect that makes computer vision, and building applications on top of it, hard. Imagine an augmented reality problem where two people look at the same scene through their phones. Assume we want to insert scaled virtual content, \eg, a virtual human, into both views. To do that in a believable fashion, we need to recover the relative pose between both cameras and we need it up to scale \cite{tateno2017cnn, arnold2022map}. 

Estimating the relative pose between two images is a long-standing problem in computer vision \cite{fathy2011fundamental, mikolajczyk2004scale, ma2021image}. Solutions based on feature matching offer outstanding quality even under adverse conditions like wide baseline matching or changing seasons \cite{jin2021image}. However, their geometric reasoning is limited to the 2D plane, so the distance between the cameras remains unknown \cite{hartley2003multiple, arnold2022map}.

In some settings, we can resort to dedicated hardware to recover the scene scale. Modern phones come with IMU sensors, but they require the user to move \cite{nutzi2011fusion}. Some phones come with LiDAR sensors that measure depth, but these sensors are limited in terms of range and constrained to very few high-end devices \cite{giubilato2018scale}.

The general setting, as recently formalized as “Map-free Relocalisation” \cite{arnold2022map}, provides only two images and intrinsics but no further measurements. The best solution to recover a metric relative pose hitherto was to combine 2D feature matching with a separate depth estimation network to lift correspondences to 3D metric space. However, there are two problems.
Firstly, the feature detector and the depth estimator are separate components that operate independently. Feature detectors generally fire on corners and depth discontinuities \cite{barroso2019key, detone2018superpoint, mikolajczyk2005comparison}, exactly the areas where depth estimators struggle.
Secondly, learning the best metric depth estimators usually requires strong supervision with ground truth depth which can be hard to come by, depending on the data domain \cite{ranftl2021vision}. \Eg, for pedestrian imagery recorded by phones, measured depth is rarely available.

We present \textbf{M}etr\textbf{ic Key}points (\ours), a feature detection pipeline that addresses both problems. Firstly, \ours regresses keypoint positions in camera space which allows us to establish metric correspondences via descriptor matching. From metric correspondences, we can recover the metric relative pose, see Figure~\ref{fig:teaser}. Secondly, by training \ours in an end-to-end fashion using differentiable pose optimization, we require only image pairs and their ground truth relative poses for supervision. Depth measurements are not required. \ours learns the correct depth of keypoints implicitly, and only for areas where features are actually found and are accurate. 
Our training procedure is robust to image pairs with unknown visual overlap, and therefore, information such as image overlap, usually acquired via structure-from-motion reconstructions \cite{li2018megadepth}, is not needed. This weak supervision makes \ours very accessible and attractive, since training it on new domains does not require any additional information beyond the poses.

\ours ranks among the top-performing methods in the Map-free Relocalization benchmark \cite{arnold2022map}, surpassing very recent, state-of-the-art approaches. \ours provides reliable, scale-metric pose estimates even under extreme viewpoint changes enabled by depth predictions that are specifically tailored towards sparse feature matching.

We summarize our \textbf{contributions} as follows: 1) A neural network, \ours, that predicts metric 3D keypoints and their descriptors from a single image, allowing metric relative pose estimation between pairs of images.
2) An end-to-end training strategy, that only requires relative pose supervision, and thus, neither depth measurements nor knowledge about image pair overlap are needed during training.

\section{Related Work}
\label{sec:related_work}

%-------------------------------------------------------------------------
\noindent\textbf{Relative Pose by Keypoint Matching.} 
In the calibrated scenario, where camera intrinsics are known, the relative camera pose can be recovered by decomposing the essential matrix \cite{hartley2003multiple}.
The essential matrix is normally computed by finding keypoint correspondences between images, and then applying a solver, \eg, the 5-point algorithm \cite{nister2004efficient}, within a RANSAC \cite{fischler1981random} loop. Classical keypoint correspondences were built around SIFT \cite{lowe1999object}, however, latest learned methods have largely superseded it. Keypoint detectors \cite{verdie2015tilde, barroso2019key, lenc2016learning}, path-based descriptors \cite{tian2019sosnet, tian2020hynet}, joint keypoint extractors \cite{detone2018superpoint, barroso2020hdd, revaud2019r2d2, dusmanu2019d2, luo2020aslfeat, gleize2023silk}, or affine shape estimators \cite{yi2016learning, mishkin2018repeatability, barroso2022scalenet} are now able to compute more accurate and distinctive features. Learned matchers \cite{sarlin2020superglue, lindenberger2023lightglue}, detector-free algorithms \cite{sun2021loftr, jiang2021cotr, truong2021learning, edstedt2023roma, chen2022aspanformer}, outlier rejection methods \cite{yi2018learning, zhao2019nm, zhang2019learning, sun2020acne, cavalli2020adalam}, or better robust estimators \cite{torr2000mlesac, chum2005matching, torr2002napsac, brachmann2019neural, barroso2023two} can improve further the quality of the estimated essential matrices.
However, the essential matrix decomposition results in a relative rotation matrix and a \emph{scaleless} translation vector. It does not yield the distance between the cameras. Arnold \etal \cite{arnold2022map} show that matching methods can resolve the scale ambiguity via metric depth estimators \cite{eigen2014depth,spencer2023slowtv}. Single-image depth prediction can regress the absolute depth in meters \cite{ranftl2021vision, liu2019planercnn, watson2019self}, being able then to lift 2D points to 3D metric coordinates, where PnP \cite{gao2003complete} or Kabsch (also called orthogonal Procrustes) \cite{kabsch1976solution, eggert1997estimating} can recover the metric relative pose from the 2D-3D or the 3D-3D correspondences, respectively.
\smalljump

\noindent\textbf{Relative Pose Regression} (RPR)
% methods compute the relative pose between a query and a reference image. Such
methods propose an alternative strategy to recover relative poses. They encode the two images within the same neural network and directly estimate their relative pose as their output. Contrary to scene coordinate regression \cite{shotton2013scene, li2020hierarchical, brachmann2018learning} or absolute pose regression \cite{kendall2015posenet, kendall2017geometric, shavit2021learning}, RPR \cite{balntas2018relocnet,cai2021extreme,winkelbauer2021learning, khatib2022grelpose,chen2021wide} does not require being trained on specific scenes, making them very versatile. 
Arnold \etal \cite{arnold2022map} propose different variants of RPR networks to tackle their AR anywhere task. 
% RPR only requires metric pose supervision during training. 
Since they do not require depth maps, Arnold \etal train their RPR networks with the supervision provided in the Map-free dataset: poses and overlap scores. 
The biggest limitation of RPR methods is that they do not provide any confidence for their estimates, making them unreliable in practice \cite{arnold2022map}. 
\smalljump

\noindent\textbf{Differentiable RANSAC}
enables optimizing pipelines that predict model parameters, \eg, camera pose parameters, directly via end-to-end training \cite{brachmann2017dsac, brachmann2019neural, wei2023generalized}. 
NG-DSAC \cite{brachmann2019neural}, a combination of DSAC  \cite{brachmann2017dsac} and NG-RANSAC \cite{brachmann2019neural}, provides gradients of RANSAC-fitted camera poses \wrt the coordinates and confidences of input correspondences.
NG-DSAC uses score function-based gradient estimates, \ie, policy gradient \cite{ding2020introduction} and REINFORCE \cite{williams1992simple}.
Besides learning visual relocalization \cite{brachmann2017dsac, brachmann2019neural}, policy gradient has also been used for learning keypoint detectors and descriptors \cite{nie2023rlsac, santellani2023s, bhowmik2020reinforced, li2022decoupling}.
$\nabla$-RANSAC \cite{wei2023generalized} is a differentiable RANSAC variation based on path derivative gradient estimators using the Gumbel softmax trick \cite{jang2016categorical}.
\smalljump

Contrary to previous keypoint extractors, \ours requires learning the 3D coordinates of keypoints. We combine NG-DSAC with the Kabsch \cite{kabsch1976solution} algorithm, a differentiable solver, and learn directly from pose signals. To the best of our knowledge, although learning from poses has been explored \cite{roessle2023end2end, wei2023generalized}, we are the first to propose a strategy that optimizes directly 3D keypoint coordinates towards metric relative pose estimation.
\begin{figure*}[hbt!]
  \centering
   \includegraphics[width=1.0\linewidth]{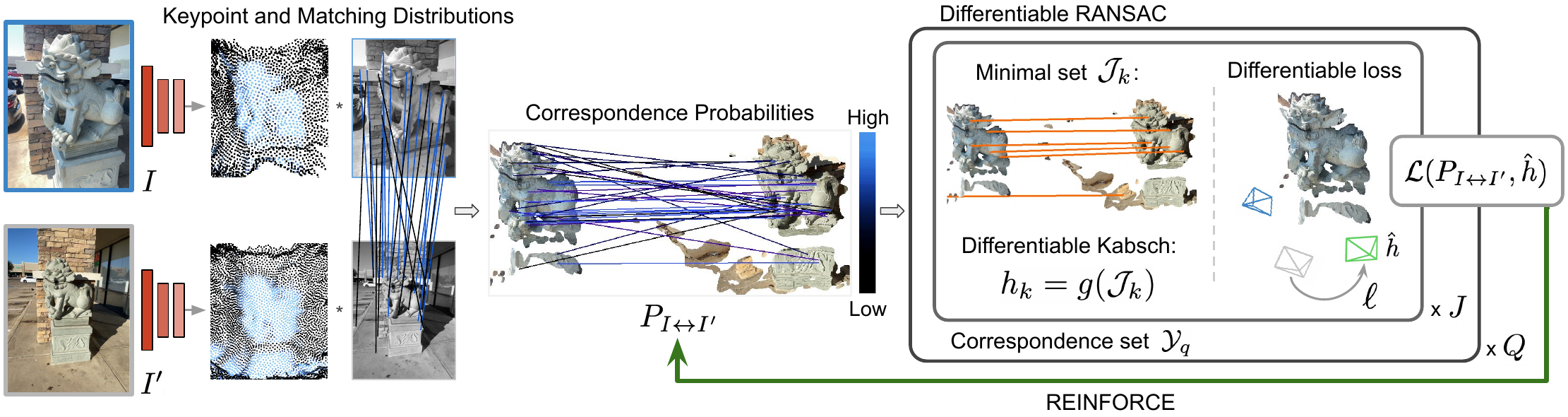}
   \vspace{-2em}
   \caption{\textbf{Training pipeline.}
   %We train \ours in an end-to-end fashion on the task of metric relative pose estimation. 
   % Some of the elements of the pipeline are not differentiable, and therefore, we redefine the task as probabilistic. Specifically, 
   \ours predicts 3D coordinates of keypoints in camera space. The network also predicts keypoint selection probabilities (keypoint distribution) and descriptors that steer the probabilities of matches (matching distribution).
   The combination of both distributions yields the probability of two keypoint being a correspondence in $P_{I \leftrightarrow I'}$, 
   and we optimize the network such that correct correspondences are more likely.
   Within a differentiable RANSAC loop, we generate multiple relative pose hypotheses and compute their loss \wrt to the ground truth transformation, $\hat{h}$.
   We generate gradients to train the correspondence probabilities $P_{I \leftrightarrow I'}$ via REINFORCE. Since our pose solver and loss function are differentiable, backpropagation also provides a direct signal to train the 3D keypoint coordinates. 
   }
   \label{fig:system}
\end{figure*}

\section{Method}
\label{sec:method}
% In this section, we explain how we compute the metric relative pose $T = (R, t)$ given \textit{I} and \textit{I'} input images.
% We first start defining the architecture and then the main components of our differentiable pose solver. 
This section first introduces the main blocks of our differentiable pose solver and then defines our new architecture.

%------------   Learning from Metric Pose Supervision   ------------
\subsection{Learning from Metric Pose Supervision}
Given two images, our system computes their metric relative pose along with the keypoint scores, the match probabilities, and the pose confidence (in the form of soft-inlier counting).
We aim to train all our relative pose estimation blocks in an end-to-end fashion. 
% Meaning that, we want to directly optimize the parameters of our network to produce accurate relative poses. 
During training, we assume to have training data as ($I$, $I'$, $T^{GT}$, $K$, $K'$), with $T^{GT}$ being the ground truth transformation and $K/K'$ the camera intrinsics. The full system is pictured in Figure \ref{fig:system}. 

To learn 3D keypoint coordinates, confidences, and descriptors, we require our system to be fully differentiable, and since some of the elements of the pipeline are not, \eg, keypoint sampling or inlier counting, we redefine the relative pose estimation pipeline as probabilistic. That means that we treat the output of our network as probabilities over potential matches, and during training, the network optimizes its outputs to generate probabilities that make correct matches more likely to be selected. 

\subsubsection{Probabilistic Correspondence Selection}
\label{sec:method:sampling}
A keypoint is characterized by a 3D coordinate, a descriptor, and a confidence value. We obtain the descriptor and the confidence directly from the output of the network, meanwhile, the 3D coordinate, $x$, is defined via a 2D position, $\bar{u}$, and a depth value, $z$: 
\vspace{-1.0em}
\begin{equation}
\begin{gathered}
    x = z \cdot K^{-1} \bar{u}^T.
    \vspace{-1.0em}
\label{eq:camera_coor}
\end{gathered}
\end{equation}
For notation simplicity, we assume homogeneous coordinates for the 2D point $\bar{u}$. 
A correspondence $y$ is defined by a keypoint in $I$ and a keypoint in $I'$, and
we refer to a set of keypoint correspondences as $\mathcal{Y}$.
We formulate the probability of drawing $\mathcal{Y}$ as the product of the probabilities of sampling them individually ($P(\mathcal{Y}) = \Pi \, P(y_{ij})$).
Specifically, we define a correspondence $y_{ij}$ as a tuple of 3D point coordinates, $y_{ij} = (x_i, x'_j)$, where $x_i$ refers to keypoint $i$ from $I$, and $x'_j$ to the keypoint $j$ from $I'$.
% In this case, $y_{ij}$ refers to the correspondence composed by keypoint $i$ and $j$ from $I$ and $I'$, respectively.
We will define the probability of sampling that correspondence next.
\smalljump

\noindent\textbf{Correspondence Probability.}
The total probability of sampling a keypoint correspondence $P(y_{ij})$ is modelled as a function of their descriptor matching and keypoint selection probabilities, such that: 
% Thus, the probability of selecting keypoint $i$ and $j$ from \textit{I} and \textit{I'} as a correspondence is given by: 
\begin{equation}
\begin{gathered}
    P(y_{ij}) = \underbrace{P_{I}(i) \cdot P_{I \rightarrow I'}(j\vert i)}_{\text{forward matching}} \cdot
    \underbrace{P_{I'}(j) \cdot P_{I \leftarrow I'}(i\vert j)}_{\text{backward matching}}.
\label{eq:corr_selection}
\end{gathered}
\end{equation}
\textit{Forward matching} combines the probability of selecting keypoint $i$ in image $I$, denoted $P_{I}(i)$, and the probability of $j$ in image $I'$ being the nearest neighbor of $i$, denoted $P_{I \rightarrow I'}(j\vert i)$. The \textit{backward matching} probability is defined accordingly.
% This probability is high when correspondences have similar and discriminative descriptors and at the same time belong to repeatable and accurate locations, while assigns low probabilities in all other cases, \eg, sky regions where there are similar descriptors but are not unique. 
We define the matrix containing all possible correspondence probabilities as $P_{I \leftrightarrow I'}$, where $P_{I \leftrightarrow I'}(i, j) = P(y_{ij})$. We formulate the descriptor matching and keypoint selection probabilities as follows:
\smalljump

\noindent\textbf{Matching Distribution.}
The descriptor matching probability represents the probability of two keypoints being mutual nearest neighbors, \ie, keypoint $i$ matches $j$, and keypoint $j$ matches $i$. For that, we first compute the probability of $j$ being the nearest neighbor to $i$ conditioned on $i$ already being selected as a keypoint: $P_{I \rightarrow I'}(j\vert i)$. We obtain that probability by applying a Softmax over all the similarities of $j$ for a fixed $i$:
\begin{equation}
\begin{gathered}
    P_{I \rightarrow I'}(j\vert i) = \text{Softmax}(m(i, \cdot)/\theta_m)_j,
\end{gathered}
\label{eq:desc_softmax_alt}
\end{equation}
where $m$ is the matrix defining all descriptor similarities and $\theta_m$ is the descriptor Softmax temperature. Before the Softmax operator, we augment $m$ with a single learnable dustbin parameter to allow unmatched keypoints to be assigned to it \cite{sarlin2020superglue}. We remove the dustbin after the Softmax operator. Equivalently, we obtain $P_{I \leftarrow I'}(i\vert j)$.
Therefore, the mutual nearest neighbor consistency, $P_{I \rightarrow I'}(j\vert i) \cdot P_{I \leftarrow I'}(i\vert j)$, is enforced by the dual-Softmax operator as in \cite{tyszkiewicz2020disk, sun2021loftr, edstedt2023dedode}.
\smalljump

\noindent\textbf{Keypoint Distribution.}
The keypoint selection probability represents the probability of two independent keypoints being sampled. \Ie, probability $P_{I}(i)$ of $i$ being selected from image $I$, and the probability $P_{I'}(j)$ of $j$ being selected from $I'$. We compute $P_{I}(i)$ by applying a spatial Softmax over all confidence values predicted by the network from image $I$, and, similarly, $P_{I'}(j)$ is obtained from the Softmax operator over all confidences from $I'$.

% We treat the score confidence maps, $C$ and $C'$, as a probability distribution over the keypoint locations. For that, we apply a spatial Softmax operator to normalize the maps independently, where each position in a map represents how likely a keypoint is to generate a good pose. Similar to the descriptor matrix, we combine $C$ and $C'$ into a single keypoint probability matrix 
% $N_{I \leftrightarrow I'}^{(w \times h) \times (w' \times h')}$
% $N_{I \leftrightarrow I'}$, where each entry represents the probability of selecting two independent keypoints from \textit{I} and \textit{I'}. 
% Note that instead of applying a local non-maxima suppression \cite{tyszkiewicz2020disk, barroso2020hdd, luo2020aslfeat}, \ours already provides a single keypoint location for every input patch.

% We finally can sample a complete set of \textit{Y} correspondences $\mathcal{Y} = \{y_{i}\}$ according to the $P_{I \leftrightarrow I'}$.

\subsubsection{Differentiable RANSAC}
Due to the probabilistic nature of our approach and potential errors in our network predictions, we rely on the robust estimator RANSAC \cite{fischler1981random} to compute pose hypotheses $h$ from $\mathcal{Y}$. We require our RANSAC layer to backpropagate the pose error to the keypoints and descriptors, and hence, we follow differentiable RANSAC works \cite{bhowmik2020reinforced, brachmann2021visual, brachmann2019neural}, and adjust them as necessary for our problem:
\smalljump

\noindent\textbf{Generate hypotheses.} Given the correspondences $\mathcal{Y}$, we use the probability values defined in $P_{I \leftrightarrow I'}$ to guide the new sampling of $J$ subsets: $\mathcal{J}_k \subset \mathcal{Y}$, with $0 \leq k < J-1$. Every subset is defined by \textit{n} 3D-3D correspondences in camera coordinates and generates a pose hypothesis $h_k$ that is recovered by a differentiable pose solver $g$, \ie, $h_k=g(\mathcal{J}_k)$.\smalljump

\noindent\textbf{Differentiable Kabsch.}
We use the Kabsch pose solver \cite{kabsch1976solution} for estimating the metric relative pose from the subset of correspondences  $\mathcal{J}_k$. 
% We obtain 3D keypoints from network outputs (as seen in Equation \ref{eq:camera_coor}) and correspondences ($\mathcal{Y}$) as explained in Section \ref{sec:method:sampling}. 
% We apply the same solver to the initial sets ($\mathcal{J}_j$) or during pose refinement ($\mathcal{I}$). 
The Kabsch solver finds the pose that minimizes the square residual over the 3D-3D input correspondences: 
\begin{equation}
\begin{gathered}
g^{\text{Kabsch}}(\mathcal{J}) = \argminB_{h} \sum_{y \in \mathcal{J}} r(y, h)^2.
\end{gathered}
\end{equation}
The residual error function $r(\cdot)$ computes the Euclidean distance between 3D keypoint correspondences after applying the pose transformation $h$. All the steps within the Kabsch solver are differentiable and have been previously studied. We refer to \cite{brachmann2021visual, avetisyan2019end} for additional details.
\smalljump

\noindent\textbf{Soft-Inlier Counting.} Since counting inliers is not differentiable, we compute a differentiable approximation of the inlier counting (soft-inlier counting) by replacing the hard threshold with a Sigmoid function $\sigma(\cdot)$. The soft-inlier counting is done over the complete set of correspondences $\mathcal{Y}$:
\begin{equation}
\begin{gathered}
    s(h, \mathcal{Y}) = \sum_{y \in \mathcal{Y}} \sigma [\beta \tau - \beta r(y, h)].
\end{gathered}
\label{eq:softInlier}
\end{equation}
% The hyper-parameter 
$\beta$ controls the softness of the Sigmoid. As in \cite{brachmann2021visual}, we set $\beta$ in dependence of the inlier threshold $\tau$ such that $\beta=\frac{5}{\tau}$.
\smalljump

\noindent\textbf{Differentiable Refinement.} We refine the pose hypothesis $h$ by iteratively finding its correspondence inliers $\mathcal{I}$ and recomputing a new pose from them:
\begin{equation}
\begin{gathered}
    h^{t+1} = g(\mathcal{I}^t) \quad  \text{and} \quad \mathcal{I}^{t+1} = \{i \vert r(y_i, h^{t+1}) < \tau\}.
\end{gathered}
\end{equation}
We repeat the process until a maximum number of iterations $t_\text{max}$ is reached, or the number of inliers stops growing. 
We refer to the refinement step as $R(h, \mathcal{Y})$, and approximate its gradients by fixing and backpropagating only through its last iteration \cite{brachmann2021visual}.

\subsubsection{Learning Objective}
\label{sec:method:loss}
We use the Virtual Correspondence Reprojection Error (VCRE) metric proposed in \cite{arnold2022map} as our loss function. VCRE defines a set of virtual 3D points ($\mathcal{V}$) and computes the Euclidean error of its projections in the image:
\begin{equation}
\begin{gathered}
\ell^{\text{VCRE}}(h, \hat{h}) =\frac{1}{\vert\mathcal{V}\vert} \sum_{v \in \mathcal{V}}\vert\vert\pi(v) - \pi(h\hat{h}^{-1}v)\vert\vert_2,
\label{eq:vcre}
\end{gathered}
\end{equation}
with $\hat{h}=T^{GT}$ and $\pi$ being the projection function. Refer to 
% supplementary material 
Section \ref{supp:training_details}
for more details on VCRE.  
For each $\mathcal{Y}$, we compute a set of hypotheses $h_k$ with corresponding scores $s_k$, and define its loss following DSAC \cite{brachmann2017dsac}:
\begin{equation}
\begin{gathered}
\ell({\mathcal{Y}}, \hat{h}) = \E_{k  \sim p(k \vert \mathcal{Y})} [\ell^{\text{VCRE}}(R(h_k, \mathcal{Y}), \hat{h})].
\label{eq:loss_Y}
\end{gathered}
\end{equation}
We derive the probability of each hypothesis, $p(k \vert \mathcal{Y})$, from its score $s_k$ via Softmax normalization.
Since the expectation above is defined over a finite set of hypotheses, we can solve it exactly to yield a single loss value for 
% a set of correspondences 
$\mathcal{Y}$.

Our final optimization is formulated as a second, outer expectation of the VCRE loss by sampling correspondence sets $\mathcal{Y}_q$ from the correspondence matrix $P_{I \leftrightarrow I'}$:
\begin{equation}
\begin{gathered}
\mathcal{L}(P_{I \leftrightarrow I'}, \hat{h}) = \E_{q  \sim p(q \vert P_{I \leftrightarrow I'})} [\ell({\mathcal{Y}_q}, \hat{h})].
\end{gathered}
\end{equation}
From now on, we abbreviate $\mathcal{L}(P_{I \leftrightarrow I'}, \hat{h})$ to $\mathcal{L}$ and $\ell({\mathcal{Y}_q}, \hat{h})$ to $\ell$. We can approximate the gradients of $\mathcal{L}$ \wrt to network parameters ($w$) by drawing $Q$ samples: 
\begin{equation}
\begin{gathered}
\frac{\partial}{\partial w} \mathcal{L} \approx \frac{1}{Q} \sum_{q \in Q} [\ell \frac{\partial}{\partial w} \text{log} P_{I \leftrightarrow I'} + \frac{\partial}{\partial w}\ell],
\end{gathered}
\label{eq:gradients_wo_baseline}
\end{equation}
where the first term provides gradients to learn descriptors and keypoint confidences, steering the sampling probability $P_{I \leftrightarrow I'}$ in a good direction.
The second term provides directly the gradients to optimize the 2D keypoint offsets and the depth estimations. 
To reduce the variance of Equation~\ref{eq:gradients_wo_baseline}, we follow \cite{brachmann2019neural} by subtracting a baseline $b = \bar{\ell}$ from $\ell$, the mean loss over all samples:
\begin{equation}
\begin{gathered}
\frac{\partial}{\partial w} \mathcal{L} \approx \frac{1}{Q} \sum_{q \in Q} [[\ell-b] \frac{\partial}{\partial w} \text{log} P_{I \leftrightarrow I'} + \frac{\partial}{\partial w}\ell].
\end{gathered}
\end{equation}

% \axel{or $b=\bar{\ell}$?}

%%------------   Curriculum Learning   ------------
\subsubsection{Curriculum Learning}
\label{sec:curr_learning}
Initialization is an important and challenging step when learning only from poses, since networks might not be able to converge without it \cite{bhowmik2020reinforced}. 
One common solution in reinforcement learning is to feed the network with increasingly harder examples that may otherwise be too difficult to learn from scratch \cite{narvekar2020curriculum}.
However, contrary to other methods \cite{sarlin2020superglue, sun2021loftr, edstedt2023roma}, we aim at training our network without requiring to know what \textit{easy} or \textit{hard} examples are. 
%Assuming a batch $B$, we define the hardest example as the image pair with the highest loss. 
Instead of using all image pairs in a training batch $B$, we optimize the network only using a subset of image pairs. We select the image pairs that return the lowest losses and ignore all others, \ie, the network is optimized only with the examples that it can solve better, and hence, are \textit{easier}.
As we progress in the training, we increase the fraction of examples from $B$ that the network needs to solve. We define both, the minimum and maximum number of pairs ($b_{\text{min}}$ and $b_{\text{max}}$), used in training.  

Even though we limit the optimization to the best pose estimates, the network might still try to optimize an image pair where all RANSAC hypotheses are incorrect.
Since hypotheses scores are normalized per image pair, see Equation~\ref{eq:loss_Y}, incorrect hypotheses can add noise to the optimization.
To mitigate this, we add a \textit{null hypothesis} ($h^0$) to Equation~\ref{eq:loss_Y}. 
The null hypothesis has a fixed score $s^0$ as well as a fixed loss $\ell^{\text{VCRE}}(h^0, \hat{h})=\text{VCRE}^{\text{max}}$, where $\text{VCRE}^\text{max}$ is the maximum value we tolerate as a \textit{good} pose. $h^0$ serves as an anchor point, such that the network can assign lower scores than $s^0$ to hypotheses with high error. 
If all hypotheses in a pool are incorrect with low scores, the null hypothesis will dominate the expectation in Equation~\ref{eq:loss_Y} and reduce the impact of gradients from the remaining hypotheses. 
% This simple yet effective strategy allows training \ours from scratch without further supervision than the poses. 

%%------------   Architecture   ------------
\subsection{Architecture}
\label{sec:method:arch}
\begin{figure}[t]
  \centering
   \includegraphics[width=1.0\linewidth]{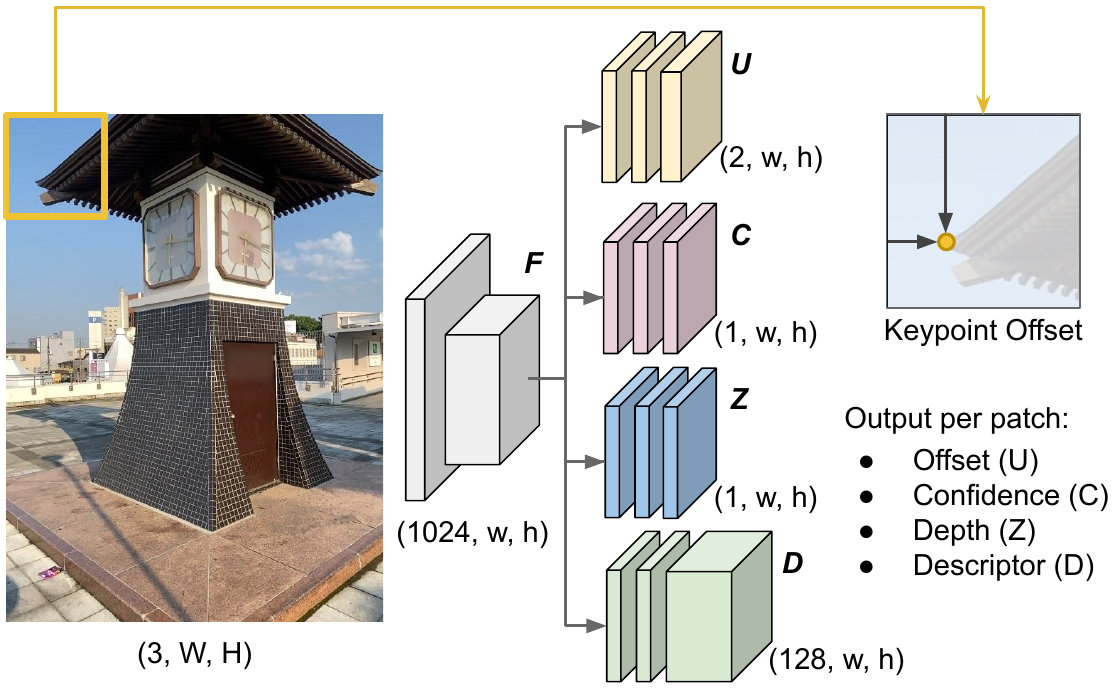}
   \caption{\textbf{\ours Architecture.} \ours uses a feature extractor that splits the image into patches. For every patch, \ours computes a 2D offset, a keypoint confidence, a depth value, and a descriptor vector. 
   % The absolute 2D position is obtained by the grid location and the 2D offset predicted, as seen in the patch example.  
   The 3D keypoint coordinates are obtained by the absolute position of the patch, its 2D offset, and depth value.
   }
   \label{fig:architecture}
\end{figure}
\ours follows a multi-head network architecture with a shared encoder \cite{tyszkiewicz2020disk, revaud2019r2d2, detone2018superpoint} that infers 3D metric keypoints and descriptors from an input image
% We show the architecture in Figure \ref{fig:architecture}. 
as seen in Figure \ref{fig:architecture}. 
\smalljump

\noindent\textbf{Encoder.} We adopt a pre-trained DINOv2 as our feature extractor \cite{oquab2023dinov2} and use its features without further training or fine-tuning. DINOv2 divides the input image into patches of size $14 \times 14$ and provides a feature vector for each patch. The final feature map, $F$, has a resolution of $(1024, w, h)$, where $w=W/14$ and $h=H/14$.
\smalljump

\noindent\textbf{Keypoint Heads.} 
We define four parallel heads that process the feature map $F$ and compute the \textit{xy} offset ($U$), depth ($Z$), confidence ($C$), and descriptor ($D$) maps; where each entry of the maps corresponds to a $14 \times 14$ patch from the input image. 
Similar to \cite{christiansen2019unsuperpoint}, \ours has the rare property of predicting keypoints as relative offsets to a sparse regular grid. We obtain the absolute 2D coordinates ($\bar{U}$) as: 
% by adding the predicted offset to its row/column position and then multiplying them by the downsampling factor $f=14$:
\begin{equation}
\begin{gathered}
    \bar{u}_{ij} = f * [u_i + i, u_j + j], \text{with } \xspace u \in [0, 1],
\label{eq:abs_coordinates}
\end{gathered}
\end{equation}
where \textit{f} refers to the encoder downsampling factor ($f=14$) and $ij$ to the grid position. 
% We then can lift keypoints to camera coordinates by combining the $\bar{C}$ 2D locations and \textit{Z} depth values together with the camera intrinsics \textit{K} as $X_{ij} = Z_{ij} K^{-1} \bar{C_{ij}}$. For simplicity, we ifnore homogenoeus coordinates. 
Since \ours predicts coordinate offsets over a coarse grid, it remains efficient while still providing correspondences with sub-pixel accuracy. 
Note that instead of applying a local non-maxima suppression \cite{tyszkiewicz2020disk, luo2020aslfeat}, \ours already provides a single keypoint location for every input patch.

\begin{table*}[t]
\footnotesize
\begin{center}
\begin{tabular}{l l c l c c l c c l c l c}
\multicolumn{13}{c}{\textbf{Map-free Dataset}}\\ 
\cline{1-13}
\noalign{\smallskip}
\cline{1-13}
\noalign{\smallskip}
\multicolumn{4}{c}{} & \multicolumn{2}{c}{VCRE} & \multicolumn{1}{c}{} & \multicolumn{2}{c}{Median Errors} & \multicolumn{1}{c}{} & \multicolumn{1}{c}{Matching} & \multicolumn{1}{c}{} & \multicolumn{1}{c}{Estimates}\\ 
\cline{5-6} \cline{8-9}  \cline{11-11} \cline{13-13} \noalign{\smallskip}
\multicolumn{4}{c}{} & \multicolumn{1}{c}{AUC} & \multicolumn{1}{c}{Prec. (\%)} & \multicolumn{1}{c}{} & \multicolumn{1}{c}{Rep. (px)} & \multicolumn{1}{c}{Trans. (m) / Rot. (\textdegree)} & \multicolumn{1}{c}{} & \multicolumn{1}{c}{Time (ms)} & \multicolumn{1}{c}{}  & \multicolumn{1}{c}{(\%)} \\ 
\cline{1-13}
\noalign{\smallskip}
&& SIFT~\cite{lowe1999object} && 0.50 & 25.0 && 222.8 & 2.93 / 61.4 && 157.6 && \textbf{100.0} \\
&& SiLK~\cite{gleize2023silk} && 0.31 & 18.0 && 176.4 & 2.20 / 33.8 && 58.4 && 52.1\\ 
\cline{1-13}
\noalign{\smallskip}
% \textbf{Depth Supervision} &&& \\
\multirow{10}{*}{\rotText{\rotatebox[origin=c]{90}{\textbf{Depth + Overlap + Pose}}}}
\multirow{10}{*}{\rotText{\rotatebox[origin=c]{90}{\textbf{Supervision}}}}
&& \textbf{Sparse Features} &&& \\
\cline{3-3} \noalign{\smallskip}
&& DISK~\cite{tyszkiewicz2020disk} && 0.54 & 26.8 && 208.1 & 2.59 / 51.9 && 58.7 && \underline{99.9} \\
&& DeDoDe~\cite{edstedt2023dedode}  && 0.53 & 31.2 && 167.4 & 2.02 / 30.2 && 200.3 && 88.0 \\
&& SuperPoint~\cite{detone2018superpoint} - SuperGlue~\cite{sarlin2020superglue}   && 0.60 & 36.1 && 160.3 & 1.88 / 25.4 && 95.6 && \textbf{100.0} \\
&& DISK~\cite{tyszkiewicz2020disk} - LightGlue~\cite{lindenberger2023lightglue} && 0.53 & 33.2 && 138.8 & \underline{1.44} / \underline{18.5} && 108.5 && 77.9 \\
\noalign{\smallskip}
&& \textbf{Dense Features} &&& \\
\cline{3-3} \noalign{\smallskip}
&& LoFTR~\cite{sun2021loftr}  && 0.61 &34.7 && 167.6 & 1.98 / 30.5 && 114.9 && \textbf{100.0} \\
&& ASpanFormer~\cite{chen2022aspanformer} && 0.64 & 36.9 && 161.7 & 1.90 / 29.2 && 177.1 && \underline{99.9} \\
&& RoMa~\cite{edstedt2023roma} && 0.67 & \underline{45.6} && \underline{128.8} & \textbf{1.23} / \textbf{11.1} && 820.2 && \underline{99.9} \\
\cline{1-13}
\noalign{\smallskip}
\multirow{9}{*}{\rotText{\rotatebox[origin=c]{90}{\textbf{No Depth}}}}
\multirow{9}{*}{\rotText{\rotatebox[origin=c]{90}{\textbf{Supervision}}}}
&& \textbf{Overlap + Pose Supervision} &&& \\
\cline{3-3} \noalign{\smallskip}
&& RPR [R($\alpha, \beta, \gamma$) + s$\cdot$t($\theta, \omega$)]~\cite{arnold2022map} && 0.35 & 35.4 && 166.3 & 1.83 / 23.2 && \textbf{21.6} && \textbf{100.0} \\
&& RPR [3D-3D]~\cite{arnold2022map} && 0.39 & 38.7 && 148.7 & 1.69 / 22.9 && 25.3 && \textbf{100.0} \\
&& RPR [R(6D) + t]~\cite{arnold2022map} && 0.40 & 40.2 && 147.1 & 1.68 / 22.5 && \underline{24.3} && \textbf{100.0 }\\
&& \textbf{MicKey w/ Overlap}  (ours) && \textbf{0.75} & \textbf{49.2} && 129.4 & 1.65 / 27.2 && 119.8 && \textbf{100.0} \\
% && MicKey w/ Overlap - Easy && 0.70 & 42.1 && 141.3 & 2.08 / 35.5 && 119.8 & 122.4 && 100.0 \\
\noalign{\smallskip}
\cdashline{2-13}\noalign{\smallskip}
&& \textbf{Pose Supervision} &&& \\
\cline{3-3} \noalign{\smallskip}
&& RPR [R(6D) + t] w/o Overlap && 0.18 & 18.1 && 197.1 & 2.45 / 34.7 && \underline{24.3} && \textbf{100.0} \\
&& \textbf{MicKey} (ours) && \underline{0.74} & \textbf{49.2} && \textbf{126.9} & 1.59 / 25.9 && 119.8 && \textbf{100.0} \\
\end{tabular}
\end{center}
\vspace{-2em}
\normalsize
\caption{\textbf{Relative pose evaluation on the Map-free dataset}. 
We report the area under the curve (AUC) and precision (Prec.) values for the VCRE metric under a threshold of 90 pixels as in \cite{arnold2022map}, where both versions of \ours obtain the top results. 
Besides, we also report the median errors, and while \ours obtains the lowest value in terms of VCRE error, other methods, \eg, RoMa, provide lower pose errors.
To compute the median errors, the benchmark only uses the valid poses generated by each method, and hence, we report the total number of estimated poses.
Lastly, we report the matching times and see that \ours is at par with LoFTR and LighGlue, while reducing significantly the time of RoMa, its closest competitor in terms of VCRE metrics. 
Matching methods use DPT \cite{ranftl2021vision} to recover the scale. 
}
\label{tab:mapfree_main_table}
\end{table*}
\section{Experiments}
\label{sec:experiments}
This section first presents details of our implementation and training pipeline, and then discusses the results for different methods in the task of instant AR at new locations. 
\smalljump 

\noindent\textbf{Inference vs Training.} 
We use the same probabilistic solver in training and testing, however, some of its parameters change. During training, given the set of correspondences $\mathcal{Y}$, we perform 20 RANSAC iterations ($J=20$), and in each one, we sample 5 correspondences ($n=5)$. Although Kabsch only requires a minimal set of 3 correspondences, we found more stable gradients when increasing it. In training, we refine all generated hypotheses. At test time, we increase the number of iterations to $J=100$ and use minimal sets of size $n=3$ in Kabsch. Moreover, we select the best hypothesis based on the soft-inlier score and only refine the winning one. 
The number of maximum refinement iterations $t_\text{max}=4$ and the number of $\mathcal{Y}$ samplings $Q=20$ is the same in training and testing.
% Refer to the supplementary for more details about the training pipeline. 
Refer to Section \ref{supp:training_details} for more details about the training pipeline. 
\smalljump 

\noindent\textbf{Map-free Benchmark} \cite{arnold2022map} evaluates the ability of methods to allow AR experiences at new locations. This task-oriented benchmark uses two images, the reference, and the query, and determines whether their estimated metric relative pose is acceptable or not for AR. A pose is accepted as \textit{good} if their VCRE, see Section \ref{sec:method:loss}, is below a threshold. Specifically, the authors argue that an offset of 10\% of the image diagonal would provide a good AR experience. In the Map-free dataset, this corresponds to 90 pixels. We follow the protocol from Map-free in two datasets; Map-free for outdoor, and ScanNet \cite{dai2017scannet} for indoor scenes. 

\subsection{Map-free Dataset} 
\begin{figure*}[t]
  \centering
   \includegraphics[width=1.0\linewidth]{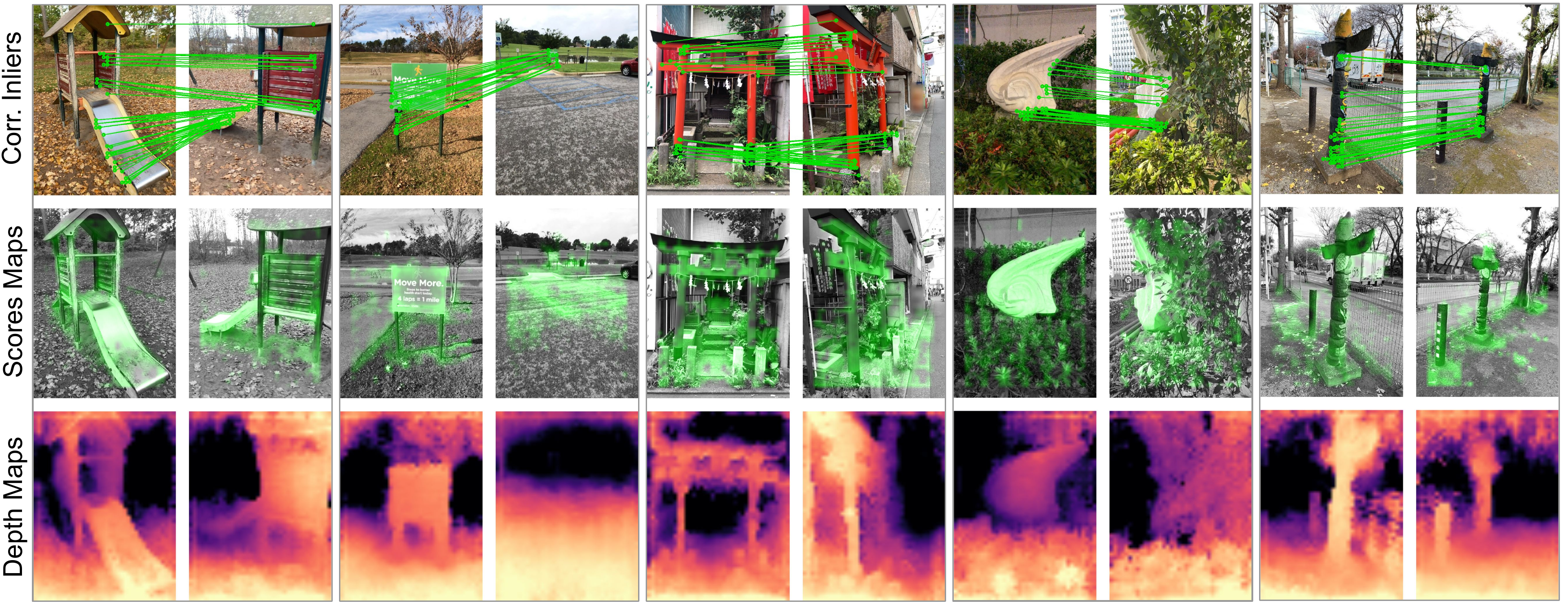}
   \caption{\textbf{Example of correspondences, scores and depth maps generated by \ours.} 
   \ours finds valid correspondences even under large-scale changes or wide baselines.
   Note that the depth maps have a resolution 14 times smaller than the input images due to our feature encoder. 
   % Even though the large scale changes or wide baseline between the images, \ours is able to find valid correspondences. 
   We follow the visualization of depth maps used in DPT \cite{ranftl2021vision}, where brighter means closer. 
   }
   \label{fig:depths_scores}   
\end{figure*}

% \begin{figure}[t]
%   \centering
%    \includegraphics[width=1.0\linewidth]{images/im_scores_depths_one_columns.pdf}
%    \caption{\textbf{Example of correspondences, scores and depth maps generated by \ours.} Note that the depth maps have a resolution 14 times smaller than the input images due to our feature encoder. 
%    % \ours finds valid correspondences even under large scale changes, 180\textdegree camera changes, or wide baseline images.
%    % Even though the large scale changes or wide baseline between the images, \ours is able to find valid correspondences. 
%    We use the same visualization for depth maps as in DPT \cite{ranftl2021vision}, where brighter means closer. 
%    }
%    \label{fig:depths_scores}   
% \end{figure}
% We first train and test \ours with the Map-free dataset. 
Map-free dataset contains 460, 65, and 130 scenes for training, validation, and test, respectively. Each training scene is composed of two different scans of the scene, where absolute poses are available. In the validation and test set, data is limited to a reference image and a sequence of query images. The test ground truth is not available, and hence, all results are evaluated through the Map-free website.  
We compare \ours against different feature matching pipelines and relative pose regressors (RPR). All matching algorithms are paired with DPT \cite{ranftl2021vision} for recovering the metric scale. 
Besides, we present two versions of \ours, one that relies on the overlap score and uses the whole batch during training, and another that follows our curriculum learning strategy (Section \ref{sec:curr_learning}). For \ours w/ Overlap, we use the same overlap range proposed in \cite{arnold2022map} (40\%-80\%). 
Evaluation in the Map-free test set is shown in Table \ref{tab:mapfree_main_table}.

The benchmark measures the capability of methods for an AR application, and instead of focusing on the relative pose errors, it quantifies the quality of such algorithms in terms of a reprojection error metric in the image plane (VCRE), claiming that this is more correlated with a user experience \cite{arnold2022map}. 
Specifically, the benchmark looks into the area under the curve (AUC) and precision value (Prec.). The AUC takes into account the confidence of the network, and hence, it also evaluates the capability of the methods to decide whether such estimations should be trusted. The precision measures the percentage of estimations under a threshold (90 pixels). We observe that the two variants of \ours provide the top VCRE results, both in terms of AUC and precision. 
We see little benefit from training \ours with also the overlap score supervision and claim that if such data is not available, our simple curriculum learning approach yields top performance. Besides, we note that training naively RPR methods without the overlap scores (RPR w/o Overlap) degrades considerably the performance. 

The benchmark also provides the reprojection error of the virtual correspondences in the image plane (see Equation \ref{eq:vcre}), and the standard translation (m) and rotation (\textdegree) errors of the poses \wrt the ground truth. 
% Refer to supplementary for more metrics and details. 
Refer to Section \ref{supp:additional_exp:mapfree} for more metrics and details. 
The median errors are computed only using valid poses, and thus, we report the total percentage of estimated poses. \ours gets the lowest reprojection error, while RoMa \cite{edstedt2023roma} obtains the lowest pose errors. 
% Although RoMa estimates are very precise, we show in the supplementary that its confidence values have low reliability, 
Although RoMa estimates are very precise, we show in Section \ref{supp:additional_exp:mapfree} that its confidence values have low reliability, 
not providing a good mechanism to decide whether such poses should be trusted. 
DISK \cite{tyszkiewicz2020disk} - LightGlue \cite{lindenberger2023lightglue} also provide accurate estimations, but they only provide 77.9\% of the total poses, meaning that the median errors are computed only for a subset of the test set. 

Lastly, for a new query image, we compute the time needed to obtain the keypoint correspondences and their 3D coordinates, \ie, \textit{xy} positions and depths. In the case of RPR, since they do direct pose regression, we report their inference time. \ours has comparable times \wrt other matching competitors, \eg, LoFTR or LightGlue, while reducing by 85\% the time of RoMa, the second best method.

\subsection{ScanNet Dataset}
Contrary to the Map-free evaluation, where depth or matching methods were trained in a different outdoor domain, we evaluate results on ScanNet \cite{dai2017scannet}, which has ground truth depths, overlap scores, and poses available for training. 
We use the training, validation, and test splits proposed by SuperGlue \cite{sarlin2020superglue} and used in following works \cite{arnold2022map, sun2021loftr, chen2022aspanformer}.
To recover the scale of matching methods, we use PlaneRCNN \cite{liu2019planercnn}, which was trained on ScanNet and yields higher quality metric poses 
% (see supplementary for details). 
(see Section \ref{supp:additional_exp:scannet} for more details). 

Evaluation in ScanNet test set is shown in Table \ref{tab:scannet_main_table}.
We use the same criteria as in the Map-free benchmark and evaluate the VCRE poses under the 10\% of the image diagonal. Contrary to Map-free, ScanNet test pairs ensure that input images overlap, and results show that all methods perform well under these conditions. Similar to previous experiment, we observe that \ours does not benefit much from using the overlap scores during training. 
Therefore, results show that training \ours with only pose supervision obtains comparable results to fully supervised methods, proving that state-of-the-art metric relative pose estimators can be trained with as little supervision as relative poses.

\begin{table}[t]
\footnotesize
\begin{center}
\begin{tabular}{c c c l c}
\multicolumn{5}{c}{\textbf{ScanNet Dataset}}\\ 
\cline{1-5}
\noalign{\smallskip}
\cline{1-5}
\noalign{\smallskip}
\multicolumn{1}{c}{} & \multicolumn{2}{c}{VCRE} & \multicolumn{1}{c}{} & \multicolumn{1}{c}{Median Errors}\\ 
\cline{2-3} \cline{5-5} \noalign{\smallskip}
\multicolumn{1}{c}{} & \multicolumn{1}{c}{AUC} & \multicolumn{1}{c}{Prec. (\%)} & \multicolumn{1}{c}{} & \multicolumn{1}{c}{Trans (m) / Rot (\textdegree)}\\ 
\cline{1-5}
\noalign{\smallskip}
% SIFT & 0.80 & 54.7 && 0.55 / 13.55 \\
% SiLK & x & x && x \\ 
% \cline{1-5} \noalign{\smallskip}
\textbf{D+O+P Signal} &&&& \\
\cline{1-1} \noalign{\smallskip}
SuperGlue~\cite{detone2018superpoint,sarlin2020superglue} & \underline{0.98} & 90.6 && 0.15 / 2.06 \\
LoFTR~\cite{sun2021loftr} & \textbf{0.99} & 91.3 && \underline{0.13} / 1.81 \\ 
ASpanFormer~\cite{chen2022aspanformer} & \textbf{0.99} & 90.6 && 0.14 / \underline{1.48} \\ 
RoMa~\cite{edstedt2023roma} & \textbf{0.99} & \textbf{94.4} && \textbf{0.11} / \textbf{1.47} \\ 
\cline{1-5} \noalign{\smallskip}
\textbf{O+P Signal} &&& \\
\cline{1-1} \noalign{\smallskip}
RPR [R + s·t]~\cite{arnold2022map} & 0.85 & 85.0 && 0.25 / 4.12 \\ 
RPR [3D-3D]~\cite{arnold2022map} & 0.93 & 92.8 && 0.20 / 4.05 \\ 
\textbf{MicKey-O} (ours) & \textbf{0.99} & \underline{93.7} && 0.17 / 3.58 \\ 
\cline{1-5} \noalign{\smallskip}
\textbf{Pose Signal} &&& \\
\cline{1-1} \noalign{\smallskip}
RPR [3D-3D] & 0.78 & 78.3 && 0.55 / 5.12 \\ 
\textbf{MicKey} (ours) & \textbf{0.99} & 92.8 && 0.17 / 3.64 \\ 
\end{tabular}
\end{center}
\vspace{-2em}
\normalsize
\caption{\textbf{Relative pose evaluation on ScanNet dataset}. All feature matching methods are coupled with PlaneRCNN \cite{liu2019planercnn} for recovering the metric scale. We indicate the training signal of each method as: depth (D), overlap score (O), and pose (P).
% \axel{networks still training - results will not change much}
}
\label{tab:scannet_main_table}
\end{table}

\subsubsection{Understanding \ours}
\begin{figure}[t]
  \centering
   \includegraphics[width=1.0\linewidth]{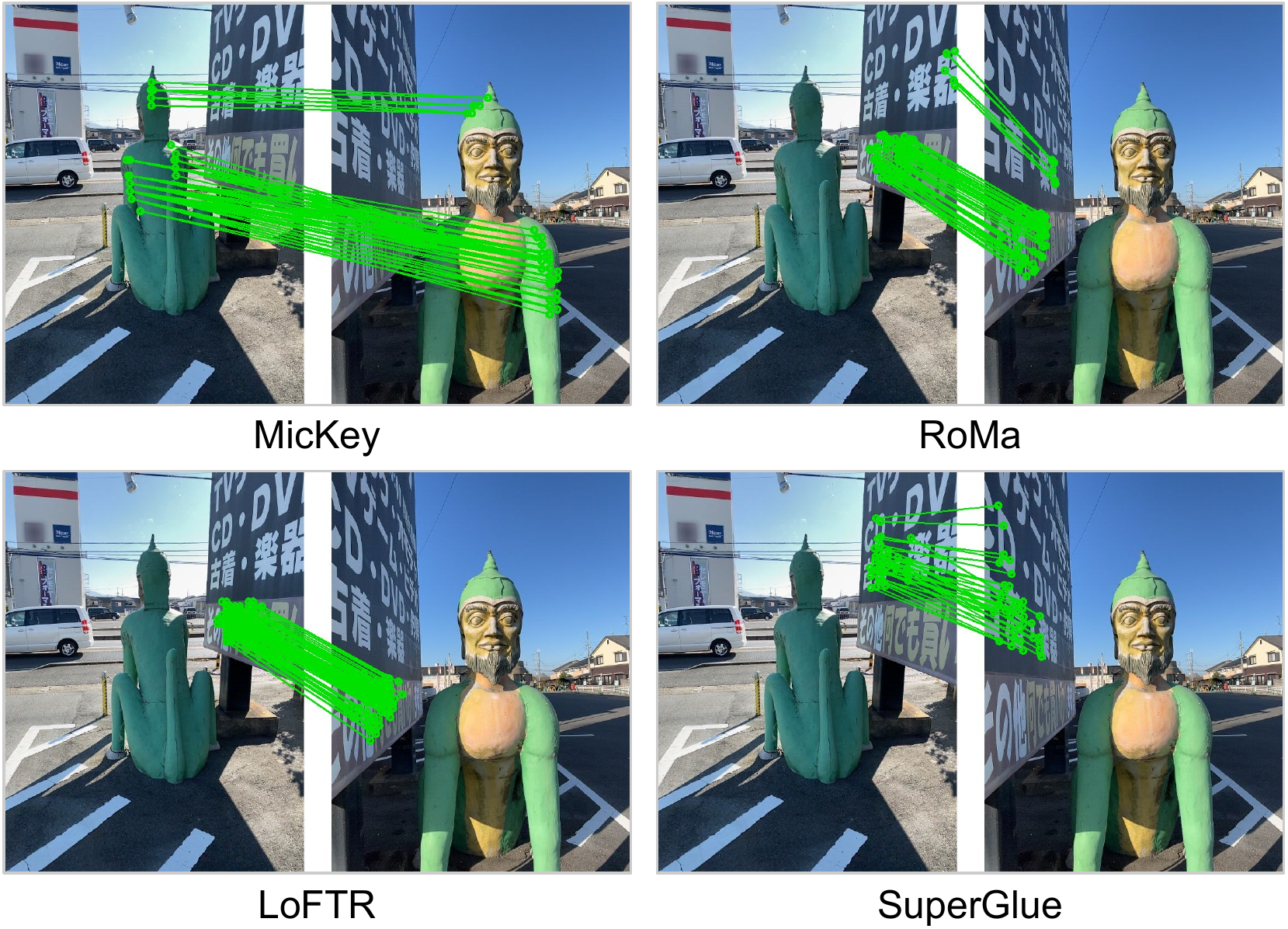}
   \vspace{-2em}
   \caption{
      \ours establishes keypoint correspondences even though images share very little visual overlap. 
      Contrary to other matchers, \ours does not focus on the textured wall, but instead reasons about the shape of the object in the foreground.
      % Contrary to other matchers, we show how \ours is able to reason about the content of the images and find metric keypoint correspondences beyond local descriptor similarities or low-level patterns. 
   }
   \label{fig:matches_comp}
\end{figure}

% \begin{table}[t]
% \footnotesize
% \begin{center}
% \begin{tabular}{cccc}
% \multicolumn{1}{c}{\textbf{AUC / Prec. (VCRE)}} & \multicolumn{1}{c}{DPT [58]} & \multicolumn{1}{c}{ZoeDepth} & \multicolumn{1}{c}{\textbf{\ours's Depth}}\\ 
% \hline \noalign{\smallskip}
% RoMa [22] & 0.67 / 45.6 & 0.67 / 47.2 & \textbf{0.73} / \textbf{52.4}\\
% \end{tabular}
% \end{center}
% \vspace{-1.8em}
% \normalsize
% \caption{\textbf{RoMa+depth} (Map-free dataset). \ours: 0.75/49.2 }
% \label{tab:ablation_roma_depth}
% % \vspace{-0.8em}
% \end{table}

% \begin{table}[t]
% \footnotesize
% \begin{center}
% \begin{tabular}{ccccc}
% \multicolumn{1}{c}{\textbf{ScanNet}} & \multicolumn{1}{c}{$\delta_1$ / $\delta_2$ / $\delta_3$ $\uparrow$} & \multicolumn{1}{c}{REL $\downarrow$} & \multicolumn{1}{c}{RMSE $\downarrow$} & \multicolumn{1}{c}{$\text{log}_{10}$ $\downarrow$} \\ 
% \hline \noalign{\smallskip}
% DPT [58] & 0.72 / 0.91 / 0.97 & 0.21 & 0.38 & 0.08 \\
% PlaneRCNN [46] & 0.75 / 0.93 / \underline{0.98} & 0.18 & 0.37 & \underline{0.07} \\
% ZoeDepth & \underline{0.79} / \underline{0.94} / \underline{0.98} & 0.17 & \textbf{0.33} & \underline{0.07} \\
% \textbf{\ours} & \underline{0.79} / \textbf{0.95} / \underline{0.98} & \underline{0.16} & 0.37 & \textbf{0.06} \\
% \textbf{\ours-Sc} & \textbf{0.80} / \textbf{0.95} / \textbf{0.99} & \textbf{0.15} & \underline{0.35} & \textbf{0.06} \\
% \end{tabular}
% \end{center}
% \vspace{-1.8em}
% \normalsize
% \caption{\textbf{Monodepth metrics} in the ScanNet dataset.}
% \label{tab:monodepth_scannet}
% % \vspace{-0.8em}
% \end{table}

\begin{table}[t]
\footnotesize
\begin{center}
\begin{tabular}{c c c l c}
\multicolumn{5}{c}{\textbf{Map-free Dataset}}\\ 
\cline{1-5}
\noalign{\smallskip}
\cline{1-5}
\noalign{\smallskip}
\multicolumn{1}{c}{} & \multicolumn{2}{c}{VCRE} & \multicolumn{1}{c}{} & \multicolumn{1}{c}{Median Errors}\\ 
\cline{2-3} \cline{5-5} \noalign{\smallskip}
\multicolumn{1}{c}{} & \multicolumn{1}{c}{AUC} & \multicolumn{1}{c}{Prec. (\%)} & \multicolumn{1}{c}{} & \multicolumn{1}{c}{Trans (m) / Rot (\textdegree)}\\ 
\cline{1-5}
\noalign{\smallskip}
% \textbf{Depth Estimation} &&& \\
% \cline{1-5} \noalign{\smallskip}
\textbf{SuperGlue}~\cite{detone2018superpoint, sarlin2020superglue} &&& \\
\cline{1-1} \noalign{\smallskip}
DPT~\cite{ranftl2021vision} & 0.60 & 36.1 && 1.88 / 25.4 \\ 
KBR~\cite{spencer2023slowtv} & 0.61 & 35.7 && 1.92 / 25.4 \\ 
\textbf{Our Depth} & \textbf{0.71} & \textbf{43.0} && \textbf{1.69} / 25.4 \\ 
\cline{1-5} \noalign{\smallskip}
\textbf{LoFTR}~\cite{sun2021loftr} &&& \\
\cline{1-1} \noalign{\smallskip}
DPT~\cite{ranftl2021vision} & 0.61 & 34.7 && 1.98 / 30.5 \\ 
KBR~\cite{spencer2023slowtv} & 0.60 & 32.1 && 2.11 / 30.5 \\ 
\textbf{Our Depth} & \textbf{0.71} & \textbf{40.7} && \textbf{1.92} / 30.5 \\ 
% \cline{1-5} \noalign{\smallskip}
% \textbf{\ours} &&& \\
% \cline{1-1} \noalign{\smallskip}
% DPT &  0.49 & 28.4 && 2.26 / 39.5 \\ 
% KBR & 0.49 & 28.3 && 2.27 / 39.7 \\ 
% Our Depth & \textbf{0.74} & \textbf{49.2} && \textbf{1.59} / \textbf{25.9}
\end{tabular}
\end{center}
\vspace{-1em}
\normalsize
\caption{\textbf{Depth ablation}. 
% Our pipeline allows for training the depth estimation head in Map-free, where only poses are available. Besides, 
Due to \ours's end-to-end nature, \ours's depth maps have been designed to provide accurate depths where keypoints are detected, boosting the performance of state-of-the-art sparse and dense feature matchers.
% benefit from using our depths. 
% \axel{MicKey w/ PnP gets 0.33 Pose AUC (LoFTR 0.35). LoFTR with MicKey Depth gets 0.47 (ROMA has 0.41)} 
% \axel{If time allows it, we could add ZoeDepth}
}
\label{tab:depth_ablation}
\end{table}
\noindent\textbf{Depth Evaluation} 
in Table \ref{tab:depth_ablation} shows that state-of-the-art matchers obtain top performance when paired with our depth maps.
% As stated in Section \ref{sec:method:arch}, \ours outputs keypoint confidence and depth maps. 
% our depths have been optimized towards sparse feature matching, and as seen in Table \ref{tab:depth_ablation}, they can be combined with different matching methods and provide top performance. 
Even though other depth methods \cite{ranftl2021vision, spencer2023slowtv} could be trained on Map-free data, it is unclear how standard photometric losses \cite{godard2017unsupervised, godard2019digging, yin2018geonet, zou2018df} would work across scans, where images could have large baselines, and whether such methods would generate better depth maps for the metric pose estimation task. 
We visualize our depth maps in Figure \ref{fig:depths_scores}, where we also display \ours's score maps and correspondence inliers. 
We see that the score maps highlight the areas of the image where the object of interest is, and it does not fire only on corners and edges. The detector and depth heads are tailored during training, and hence, the detector learns to use the positions where the depth is accurate. 
\smalljump 

% \begin{figure*}[t]
%   \centering
%    \includegraphics[width=1.0\linewidth]{images/matches_3d.pdf}
%    \caption{\textbf{Visualization of relative camera pose estimation under extreme viewpoint difference.} 
%    We use Map-free validation scenes and visualize the different predicted poses together with their ground truth (green camera). For these challenging examples, we see that \ours is the most consistent in computing a good relative pose. Only in the last example, RoMA computes a slightly more accurate pose. 
%    }
%    \label{fig:matches}
% \end{figure*}

\begin{figure}[t]
  \centering
   \includegraphics[width=1.0\linewidth]{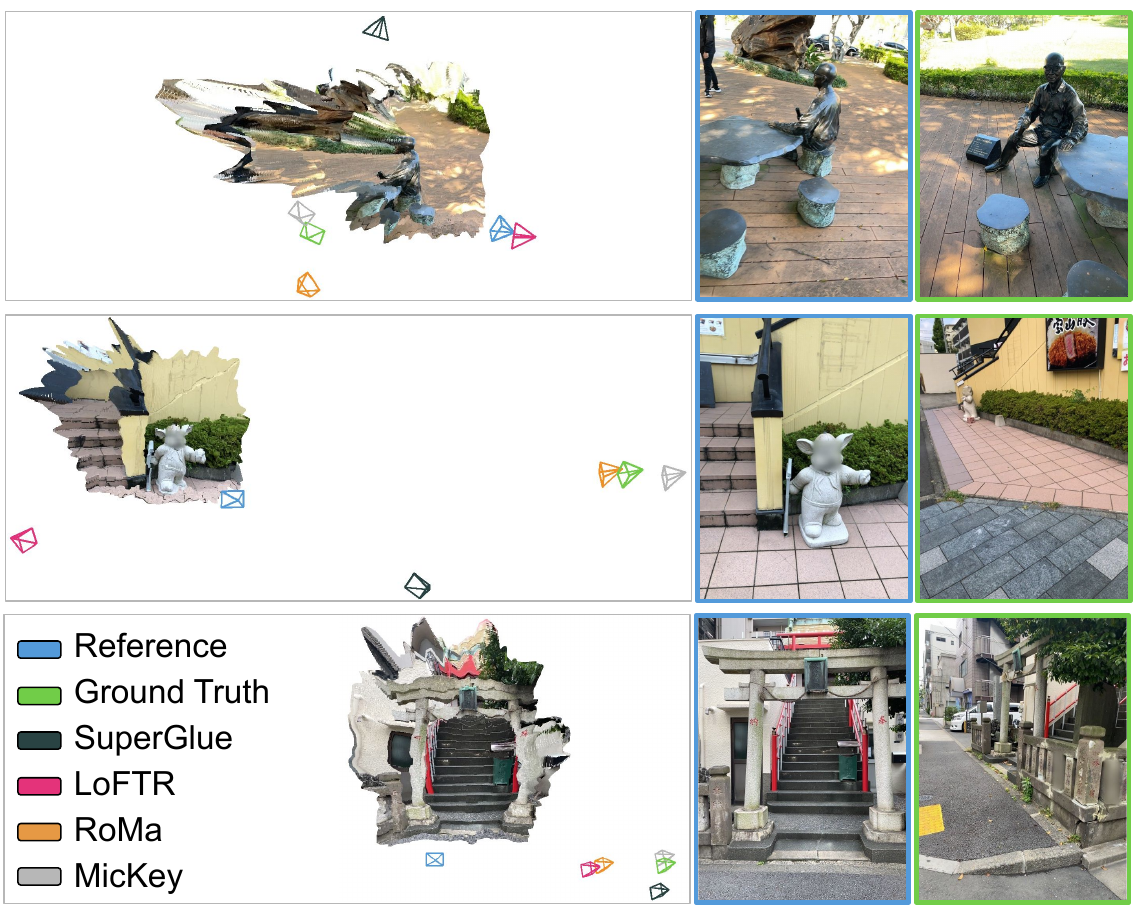}
   \caption{\textbf{Visualization of relative camera poses.}
   % estimation under extreme viewpoint difference.} 
   In this visualization, we use Map-free validation scenes and visualize the different predicted poses \wrt the ground truth (green camera). 
   % For these challenging examples, we see that \ours is the most consistent in computing a good relative pose. Only in the last example, RoMA computes a slightly more accurate pose. 
   }
   \label{fig:matches}
\end{figure}
\noindent\textbf{Beyond Low-level Matching.} Figure \ref{fig:matches_comp} displays an image pair where almost no visual overlap is shared between the images. 
In that example, we see that \ours exhibits a different behavior than the other matchers, where instead of focusing on the low-level patterns of the wall, it reasons about the shape of the object, showing glimpses of high-level reasoning.
State-of-the-art competitors have been trained with the premise that image pairs have a minimum overlap, meanwhile, our flexible training pipeline uses non-overlapping examples during training and learns to deal with them. 
Moreover, such methods use depth consistency checks for establishing ground truth correspondences, and in such cases, some of the matches computed by \ours would not be proposed as candidates.
Moreover, we 
visualize the 3D camera locations of challenging examples in Figure \ref{fig:matches}, and report results in 
% the supplementary 
Section \ref{supp:additional_exp:mapfree}
for a split of only extreme viewpoint pairs, where
we see that this behavior is not isolated, and \ours obtains the best results.\smalljump 

% when image pairs are under extreme viewpoint changes.
% highest number of frames under the VCRE threshold. 
% and see that not only looking for matches in textured areas can improve the pose estimates under extreme viewpoint changes. We refer to the supplementary for evaluation on extreme image pairs.

% We display the estimated poses in Figure \ref{fig:matches}, where we see that predicting jointly keypoints and depth values surpasses independent depth and keypoint predictors.  

% matches the object in the foreground and ignores the  
% exhibit different behaviour . and try to match the object in the foreground 
% describe examples

% Instead of matching low-level patterns, reasons about the shape of the object, showing glimpses of high level reasonigs.

% supervision issue, our advantage:
% Contrary to other methods, which focuses on low-level patterns or local descriptor similarities, \ours reasons about the image content and provides matches that go beyond state-of-the-art matcher capabilities. 

% rarely seen by other matcher due to minimum overlap score, our flexible schedule shows these examples and can deal with them. 

% correspondences/occlusion when obtaining gt corr. from depth.

% We then emphasize that the end-to-end supervision allows for high-level understanding of the task, providing correspondences that pixel level supervision would not be able to obtain.

% 

\smalljump 

\noindent\textbf{Limitations.}
As seen in Tables \ref{tab:mapfree_main_table} and \ref{tab:scannet_main_table}, \ours excels at estimating good poses within an accuracy threshold that is useful for AR applications. For very fine thresholds, other methods could obtain more accurate pose estimates, \ie, their translation and rotation errors are smaller. 
\Eg, see the second example in Figure \ref{fig:matches}.
The DINOv2 features we use are very powerful, but limited in resolution \cite{oquab2023dinov2}. Future work could investigate backbone architectures that enable higher-resolution feature maps without compromising the expressiveness of our current feature encoder.
% challenging image pairs from the validation set. 
% On those, \ours is able to find good poses for an AR experience.
% However, we see that the poses do not align with the ground truth perfectly, and in one of the examples, RoMa is able to compute a more precise pose. This underlines that there is still room for improvement, and that additional refinement or more accurate keypoint coordinates could improve the quality of the poses. 

\section{Conclusions}
\label{sec:conclusions}
We present \ours, a neural network that enables 2D image matching in 3D camera space. Our evaluation shows that \ours ranks on top of the Map-free relocalization benchmark, where only weak training supervision is available, and obtains better or comparable results to other state-of-the-art methods in ScanNet, where methods were trained with full supervision.
Thanks to our end-to-end training, we showed that \ours can compute correspondences beyond low-level pattern matching.
Besides, entangling the keypoint and depth estimation during training showed that our depth maps are tailored towards the feature matching task, and top-ranking matchers perform better with our depths.
Our experiments prove that we can train state-of-the-art keypoint and depth regressors without strong supervision.

\begin{appendices}
\section{Architecture Details}
Complementary to the details and definitions of \ours from the main sections, we include the details of its implementation in Tables \ref{sup:tab:mickey_details} and \ref{sup:tab:mickey_details_2}. As mentioned, from a single input image $I$, \ours computes the 2D keypoint offset (U), confidence (C), depth (Z), and descriptor (D) maps. The network is split into two main blocks, the feature encoder and the keypoint heads. We use as our feature encoder DINOv2, without further training or fine-tuning. We refer to its original paper \cite{oquab2023dinov2} for additional details on DINOv2. In Tables \ref{sup:tab:mickey_details} and \ref{sup:tab:mickey_details_2}, we detail the layers within the different keypoint heads. 
Each keypoint head is composed of ResNet blocks \cite{agarap2018deep}, a small self-attention layer, and specific activation functions.
As explained in the tables, a ResNet block is composed of $3\times3$ convolutions, batch normalization layers, ReLU activations, and a residual connection. The residual connection is done between the input of the block and its output. 
Given that DINOv2 is not trained, in every head, we add a small self-attention layer to allow trainable message-passing systems within the network. We use linear attention to reduce the computational complexity \cite{katharopoulos2020transformers}. The transformer has three attention layers, and each layer has eight attention heads.
Note that we do not apply the transformer right after DINOv2 encoder, but instead, we use it after processing the feature maps to a smaller descriptor dimension to reduce the overall complexity and memory of our network. 
Finally, at the end of the heads, we use different activation functions depending on the keypoint head. For instance, we apply a Sigmoid activation to the keypoint offsets to map them to the range $[0, 1]$, \ie, the offset is allowed to move within its corresponding grid.

\begin{table}[th!]
% \tiny
\footnotesize
\begin{center}
\begin{tabular}{c c c}

\multicolumn{3}{c}{\textbf{Offset Head (U)}}\\
\hline \noalign{\smallskip}
Layer & Description & Output Shape\\
\hline \noalign{\smallskip}
& Feature map $F$ & [b, 1024, w, h]\\
1 & ResNet block 1 & [b, 512, w, h]\\
2 & ResNet block 2 & [b, 256, w, h]\\
3 & ResNet block 3 & [b, 128, w, h]\\
4 & Self-Attention & [b, 128, w, h]\\
5 & ResNet block 4 & [b, 64, w, h]\\
6 & Conv. - Sigmoid & [b, 2, w, h]\\

\hline \noalign{\smallskip}
\hline \noalign{\smallskip}
% \cline{2-9} \noalign{\smallskip}
\multicolumn{3}{c}{\textbf{Depth Head (Z)}}\\
\hline \noalign{\smallskip}
Layer & Description & Output Shape\\
\hline \noalign{\smallskip}
& Feature map $F$ & [b, 1024, w, h]\\
1 & ResNet block 1 & [b, 512, w, h]\\
2 & ResNet block 2 & [b, 256, w, h]\\
3 & ResNet block 3 & [b, 128, w, h]\\
4 & Self-Attention & [b, 128, w, h]\\
5 & ResNet block 4 & [b, 64, w, h]\\
6 & $3\times3$ Convolution & [b, 1, w, h]\\

\end{tabular}
\end{center}
\normalsize
\vspace{-0.40cm}
\caption{\textbf{3D Keypoint Coordinate Heads.}
The feature extractor computes features $F$ from an input image $I$. The dimension of the feature map is $(1024, w, h)$, where $w=W/14$ and $h=H/14$, and $W$ and $H$ refer to the width and height of $I$. 
The feature map $F$ is then processed by different keypoint heads in parallel. 
A ResNet block is composed of $3\times3$ convolutions, batch normalization layers \cite{ioffe2015batch}, ReLU activations \cite{agarap2018deep}, and a residual connection. The Self-Attention layer refers to a self-attention transformer with linear attention \cite{katharopoulos2020transformers}.
}
\label{sup:tab:mickey_details}
\end{table}

\begin{table}[th!]
% \tiny
\footnotesize
\begin{center}
\begin{tabular}{c c c}
% \cline{2-9} \noalign{\smallskip}
\multicolumn{3}{c}{\textbf{Confidence Head (C)}}\\
\hline \noalign{\smallskip}
Layer & Description & Output Shape\\
\hline \noalign{\smallskip}
& Feature map $F$ & [b, 1024, w, h]\\
1 & ResNet block 1 & [b, 512, w, h]\\
2 & ResNet block 2 & [b, 256, w, h]\\
3 & ResNet block 3 & [b, 128, w, h]\\
4 & Self-Attention & [b, 128, w, h]\\
5 & ResNet block 4 & [b, 64, w, h]\\
6 & Conv. - Spatial Softmax & [b, 1, w, h]\\

\hline \noalign{\smallskip}
\hline \noalign{\smallskip}
% \cline{2-9} \noalign{\smallskip}
\multicolumn{3}{c}{\textbf{Descriptor Head (D)}}\\
\hline \noalign{\smallskip}
Layer & Description & Output Shape\\
\hline \noalign{\smallskip}
& Feature map $F$ & [b, 1024, w, h]\\
1 & ResNet block 1 & [b, 512, w, h]\\
2 & ResNet block 2 & [b, 256, w, h]\\
3 & ResNet block 3 & [b, 128, w, h]\\
4 & Self-Attention & [b, 128, w, h]\\
5 & ResNet block 4 & [b, 128, w, h]\\
6 & L2 Normalization & [b, 128, w, h]\\

\end{tabular}
\end{center}
\normalsize
\vspace{-0.40cm}
\caption{\textbf{Confidence and Descriptor Heads.}
Similar to the 3D coordinate regressors, \ours also computes the descriptors and confidence scores of each keypoint. Each head has a different activation function, \eg, in the descriptor head, we L2 normalize the descriptors and map them to a sphere of radius 1. 
}
\label{sup:tab:mickey_details_2}
\end{table}

\section{Training Details}
\label{supp:training_details}
\noindent\textbf{Training parameters.} 
We train \ours with a batch size of 48 image pairs. At the start of the training, we use only the 30\% of pairs to optimize the network ($b_{\text{min}}=14$), and linearly increase the number of used pairs by 10\% every 4k training iterations. We stop increasing the number of considered pairs when we reach the 80\% of the batch ($b_{\text{max}}=38$). The warm-up period finishes after 20k iterations. For the \textit{null hypothesis}, we define $\text{VCRE}^{\text{max}}=120$ pixels and $s^0$ is defined as the 30\% of correspondences being inliers, \ie, if having a correspondence set of size 100, $s^0=30$.
During correspondence selection, we define the temperature of the descriptor Softmax 
% (Equation 3 in the main paper) 
(Equation \ref{eq:desc_softmax_alt}) 
as $\theta_m=0.1$ and initialize the learnable dustbin parameter to 1. 
% In Equation 5 of the main paper, 
In Equation \ref{eq:softInlier}, 
$\beta$ controls the smoothness on the soft-inlier counting, and it is defined in dependence of the inlier threshold, $\tau$. We define the threshold as $\tau=0.15$m. 
\smalljump

\noindent\textbf{Optimization.} 
\ours is trained in an end-to-end manner with randomly initialized weights and ADAM optimizer \cite{kingma2014adam} with a learning rate of $10^{-4}$. We train on four V100 GPUs and the network converges after seven days.
To pick the best checkpoint, we evaluate the performance in a subset of the validation dataset in terms of the Area Under the Curve (AUC) of the VCRE metric. We check validation results twice at every epoch, which corresponds to $\sim$1k training iterations.
\smalljump

\noindent\textbf{Virtual Correspondences.} 
% Equation 7 from the main paper 
Equation \ref{eq:vcre}
uses virtual correspondences to compute the Virtual Correspondence Reprojection Error (VCRE). We define such virtual correspondences as in Map-free benchmark \cite{arnold2022map}. The virtual correspondences represent a uniform grid of 3D points that will be projected into the 2D image plane to compute the VCRE. We define a total of 196 virtual correspondences ($\vert\mathcal{V}\vert=196$). They correspond to a cube of $2.1 \times 1.2 \times 2.1$ meters in XYZ coordinates, where the minimum separation between 3D points is $0.3$m. Note that this formulation already embeds the quality of an estimated pose in a single value, the VCRE. Hence, contrary to the standard pose loss formulation, using VCRE as a loss function does not require a parameter that balances the translational and rotational components of the pose error loss~\cite{barroso2023two, roessle2023end2end}.

\section{Additional Experiments}

In addition to the experiments and visualization reported in the main paper, we provide more insights, visualizations, and experiments in this section.

\subsection{ScanNet}
\label{supp:additional_exp:scannet}
\label{sup:exp:scannet}

\begin{table}[t]
\footnotesize
\begin{center}
\begin{tabular}{c c c l c}
\multicolumn{5}{c}{\textbf{ScanNet Dataset}}\\ 
\cline{1-5}
\noalign{\smallskip}
\cline{1-5}
\noalign{\smallskip}
\multicolumn{1}{c}{} & \multicolumn{2}{c}{VCRE} & \multicolumn{1}{c}{} & \multicolumn{1}{c}{Median Errors}\\ 
\cline{2-3} \cline{5-5} \noalign{\smallskip}
\multicolumn{1}{c}{} & \multicolumn{1}{c}{AUC} & \multicolumn{1}{c}{Prec. (\%)} & \multicolumn{1}{c}{} & \multicolumn{1}{c}{Trans (m) / Rot (\textdegree)}\\ 
\cline{1-5}
\noalign{\smallskip}
\textbf{Depth Estimation} &&& \\
\cline{1-5} \noalign{\smallskip}
\textbf{SuperGlue}~\cite{detone2018superpoint,sarlin2020superglue} &&& \\
\cline{1-1} \noalign{\smallskip}
DPT~\cite{ranftl2021vision} & 0.98 & 90.0 && 0.17 / \textbf{2.06} \\ 
PlaneRCNN~\cite{liu2019planercnn} & 0.98 & 90.6 && 0.15 / \textbf{2.06} \\ 
\textbf{Our Depth} & \textbf{0.99} & \textbf{91.7} && \textbf{0.11} / \textbf{2.06} \\ 
% \smallskip 
\hdashline \noalign{\smallskip}
GT Depth & 0.99 & 92.9 && 0.07 / 2.06 \\ 
\cline{1-5} \noalign{\smallskip}
\textbf{LoFTR}~\cite{sun2021loftr} &&& \\
\cline{1-1} \noalign{\smallskip}
DPT~\cite{ranftl2021vision} & \textbf{0.99} & 89.4 && 0.16 / \textbf{1.81} \\ 
PlaneRCNN~\cite{liu2019planercnn} & \textbf{0.99} & \textbf{91.3} && 0.13 / \textbf{1.81} \\ 
\textbf{Our Depth} & \textbf{0.99} & 90.3 && \textbf{0.10} / \textbf{1.81} \\ 
% \smallskip 
\hdashline \noalign{\smallskip}
GT Depth & 0.99 & 91.3 && 0.07 / 1.81 \\ 
% \cline{1-5} \noalign{\smallskip}
% \textbf{\ours} &&& \\
% \cline{1-1} \noalign{\smallskip}
% DPT & x & x && x \\ 
% PlaneRCNN & x & x && x \\ 
% Our Depth & 0.99 & 92.8 && 0.17 / 3.64 \\ 
% \smallskip 
% \hdashline \noalign{\smallskip}
% GT Depth & x & x && x \\ 
\end{tabular}
\end{center}
\vspace{-2em}
\normalsize
\caption{\textbf{Relative pose evaluation on ScanNet for different depth estimators}.
% Our pipeline trains the depth estimation head in ScanNet from only pose supervision. 
We show here the results of different matching algorithms paired with different depth estimators and \ours's depths.
% as well as methods that have been fully supervised with the ground truth (GT) depths (\mbox{PlaneRCNN}). 
SuperGlue, a sparse feature matcher, obtains top results when combined with our depth estimations. This is in line with our system, since we optimize our depth head to work with sparse rather than dense features. 
% \axel{MicKey w/ PnP gets 0.33 Pose AUC (LoFTR 0.35). LoFTR with MicKey Depth gets 0.47 (ROMA has 0.41)} 
}
\label{tab:sup:depth_ablation}
\end{table}

\noindent\textbf{Depth ablation}. Similar to the ablation study on depth estimation methods done in the main paper, we also report the results of sparse and dense matchers combined with different depth estimators in Table \ref{tab:sup:depth_ablation}. Specifically, we use DPT~\cite{ranftl2021vision} and PlaneRCNN~\cite{liu2019planercnn}, where the latest used the ground truth depth maps provided in ScanNet for its training. As a reference, we provide  the results when combining the matchers with the ground truth depths. 
Even though \ours did not use any ground truth depth data during training, both matchers, SuperGlue~\cite{detone2018superpoint, sarlin2020superglue} and LoFTR~\cite{sun2021loftr}, benefit from using our depth maps. SuperGlue, a sparse feature method like \ours, is the one that yields better results with our depths.
\ours's depth maps were trained specifically for sparse matching, and hence, a sparse matching method could benefit more from them. 
\smalljump

\begin{table}[t]
\footnotesize
\begin{center}
\begin{tabular}{ccccc}
\multicolumn{1}{c}{\textbf{ScanNet}} & \multicolumn{1}{c}{$\delta_1$ / $\delta_2$ / $\delta_3$ $\uparrow$} & \multicolumn{1}{c}{REL $\downarrow$} & \multicolumn{1}{c}{RMSE $\downarrow$} & \multicolumn{1}{c}{$\text{log}_{10}$ $\downarrow$} \\ 
\hline \noalign{\smallskip}
DPT \cite{ranftl2021vision} & 0.72 / 0.91 / 0.97 & 0.21 & 0.38 & 0.08 \\
PlaneRCNN \cite{liu2019planercnn} & 0.75 / 0.93 / \underline{0.98} & 0.18 & 0.37 & \underline{0.07} \\
ZoeDepth \cite{bhat2023zoedepth} & \underline{0.79} / \underline{0.94} / \underline{0.98} & 0.17 & \textbf{0.33} & \underline{0.07} \\
\textbf{\ours} & \underline{0.79} / \textbf{0.95} / \underline{0.98} & \underline{0.16} & 0.37 & \textbf{0.06} \\
\textbf{\ours-Sc} & \textbf{0.80} / \textbf{0.95} / \textbf{0.99} & \textbf{0.15} & \underline{0.35} & \textbf{0.06} \\
\end{tabular}
\end{center}
\vspace{-2.0em}
\normalsize
\caption{\textbf{Indoor monocular depth evaluation.} Even though \ours depth maps are optimized to produce precise relative poses, and hence, might not be accurate beyond keypoint positions, we see that \ours still produces accurate depth maps that are on par with or surpass current state-of-the-art methods.}
\label{tab:sup:monodepth_scannet}
\vspace{-0.7em}
\end{table}
\begin{figure*}[t]
  \centering
   \includegraphics[width=1.0\linewidth]{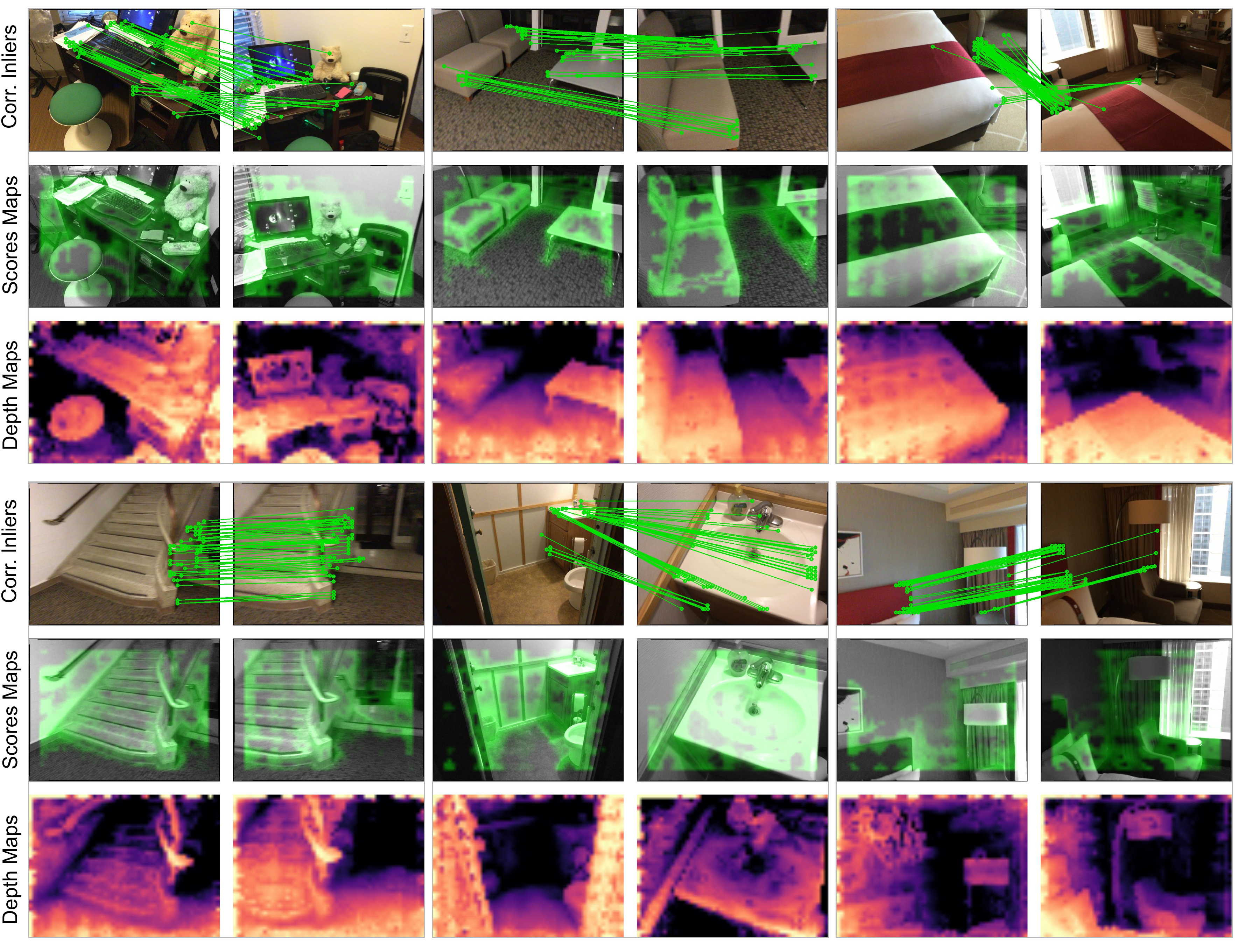}

   \caption{\textbf{Visual examples on ScanNet dataset.} 
   We visualize the inlier correspondences, score, and depth maps predicted by \ours in ScanNet images. We see that although the images show repetitive patterns and flat areas, \ours is able to find correct correspondences across views. Besides making our correspondences metric, our depth estimation head provides additional 3D information and geometric constraints to our probabilistic pose solver, making the final decision robust against the mentioned indoor challenges.    
   }
   \label{fig:sup:scannet_visual}
\end{figure*}

\noindent\textbf{Monocular depth evaluation}. Table \ref{tab:sup:monodepth_scannet} presents monocular depth evaluation metrics for the ScanNet test set, where we have access to GT depths. For completeness, besides DPT \cite{ranftl2021vision} and PlaneRCNN \cite{liu2019planercnn}, we also evaluate the depth estimations of recent ZoeDepth \cite{bhat2023zoedepth}. 
\ours-Sc corresponds to evaluating the depth estimates only on the positions where \ours is most confident. Specifically, we take the depth estimations that correspond to the top 50\% scoring positions. And therefore, in this setup, we focus on the positions that will be used to compute the metric relative pose between images. We see that our relative pose supervision, even though not using any depth maps during training, allows \ours to compute accurate depth maps.

\smalljump
\noindent\textbf{Visual examples}. We also show some visual examples of the depth maps estimated by \ours in Figure \ref{fig:sup:scannet_visual}. Moreover, in the figure, we display the inlier correspondences and the score maps that \ours computes for each input image. We see that \ours finds high scoring keypoints in structures beyond corners or blobs, establishing correct matches in images where there are few discriminative structures.

\subsection{Map-free}
\label{supp:additional_exp:mapfree}
\begin{figure}[t]
  \centering
   \includegraphics[width=1.\linewidth]{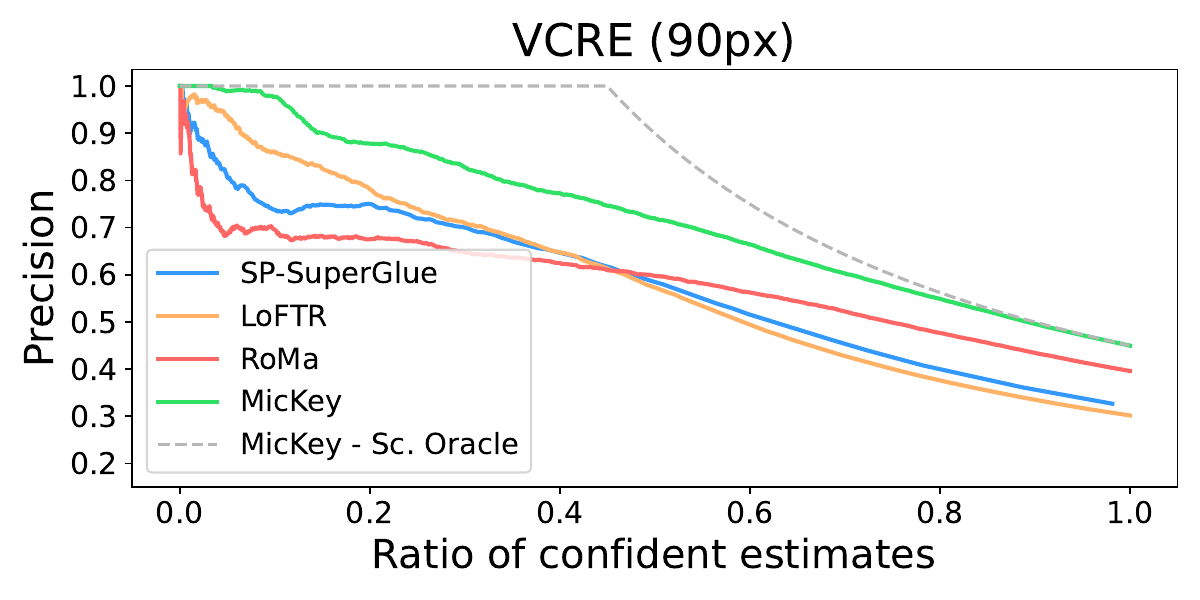}
   \vspace{-2em}
   \caption{\textbf{Precision vs ratio of estimates.}
   In this plot, the pose estimates are ranked by the confidence values of the methods, \eg, inlier counting. The Map-free benchmark computes the AUC of the curves to also evaluate the ability of the methods to decide whether their poses should be trusted or not. 
   We plot \ours - \mbox{Score (Sc.) Oracle}, which corresponds to the curve that \ours would obtain if its pose confidences were perfect.
   }
   \label{fig:auc_vcre}
\end{figure}
\noindent\textbf{Method Confidence} tells when we can or not trust a pose estimate. Map-free benchmark evaluates the confidence of the methods via the area under the curve (AUC) metric. The benchmark ranks the poses by confidence, and hence, the AUC is only maximized when the most accurate poses are ranked first. In Figure \ref{fig:auc_vcre}, we visualize the precision versus the ratio of estimates in the validation set scenes, where ground truth poses are available. 
All matching methods use their inlier counting as their confidence value, meanwhile, \ours relies on the soft-inlier counting. 
% RPR does not provide confidence and hence, 
\ours's confidence, hence, is entangled within its training pipeline and directly optimized to be correlated with the quality of its pose predictions. From Figure \ref{fig:auc_vcre}, we see that \ours obtains the highest number of correctly ranked images before 
assigning a high score to an invalid pose, where an invalid pose refers to a relative pose with a VCRE higher than 90 pixels. 
Contrary, as seen in the plot, the second best method, RoMa \cite{edstedt2023roma}, struggles to rank its pose estimates. \Ie, RoMa accepts incorrect poses (VCRE \textgreater{} 90px) as its most confidence estimates.  This indicates that RoMa's poses, although very accurate, do not have a valid mechanism to decide whether they should be trusted or rejected, making them unreliable in an AR application \cite{arnold2022map}. 
\smalljump

\begin{table}[t]
\footnotesize
\begin{center}
\begin{tabular}{c l c c c c}
\multicolumn{6}{c}{\textbf{Map-free Dataset}}\\ 
\cline{1-6}
\noalign{\smallskip}
\cline{1-6}
\noalign{\smallskip}
\multicolumn{2}{c}{} & \multicolumn{2}{c}{VCRE (90px)} & \multicolumn{2}{c}{Pose (25cm, 5\textdegree)}\\ 
\cline{3-4} \cline{5-6} \noalign{\smallskip}
\multicolumn{2}{c}{} & \multicolumn{1}{c}{AUC} & \multicolumn{1}{c}{Prec. (\%)} & \multicolumn{1}{c}{AUC} & \multicolumn{1}{c}{Prec. (\%)}\\ 
\cline{1-6}
\noalign{\smallskip}
\textbf{D+O+P Signal} &&&& \\
\cline{1-1} \noalign{\smallskip}
SuperGlue \cite{detone2018superpoint, sarlin2020superglue} && 0.60 & 36.1 & 0.35 & \underline{16.8} \\
LightGlue \cite{tyszkiewicz2020disk, lindenberger2023lightglue} && 0.53 & 33.2 & 0.31 & 15.8 \\
DeDoDe \cite{edstedt2023dedode} && 0.53 & 31.2 & 0.26 & 12.5 \\
LoFTR \cite{sun2021loftr} && 0.61 & 34.7 & 0.35 & 15.4 \\
ASpanFormer \cite{chen2022aspanformer} && 0.64 & 36.9 & \underline{0.36} & 16.3 \\
RoMa \cite{edstedt2023roma} && 0.67 & \underline{45.6} & \textbf{0.41} & \textbf{22.8} \\
\cline{1-6} \noalign{\smallskip}
\textbf{O+P Signal} &&& \\
\cline{1-1} \noalign{\smallskip}
RPR [R(6D) + t] \cite{arnold2022map} && 0.40 & 40.2 & 0.06 & 6.0 \\
\textbf{MicKey-O} (ours) && \textbf{0.75} & \textbf{49.2} & 0.33 & 13.3 \\
\cline{1-6} \noalign{\smallskip}
\textbf{Pose Signal} &&& \\
\cline{1-1} \noalign{\smallskip}
RPR [R(6D) + t] && 0.18 & 18.1 & 0.01 & 0.6 \\
\textbf{MicKey} (ours) && \underline{0.74} & \textbf{49.2} & 0.28 & 12.0 \\
\end{tabular}
\end{center}
\vspace{-2em}
\normalsize
\caption{\textbf{Additional metrics on Map-free dataset.}
Besides the VCRE results, we also show the AUC and precision values for a very fine threshold (pose errors at 25cm and 5\textdegree).
}
\label{tab:sup:mapfree_additional}
\end{table}
\noindent\textbf{Pose Metrics.}
Besides the experiments from the main paper in Map-free, we provide additional metrics in Table \ref{tab:sup:mapfree_additional}. Map-free benchmark, although it focuses on the VCRE metric for evaluating algorithms for an AR experience, it also computes the AUC and precision error pose under a very fine threshold (pose error \textless{} 25cm, 5\textdegree{}). Under such conditions, all methods have a small AUC and precision value, and hence, applications built on that restrictive threshold would need to discard most of the relative pose estimates. Even though, it gives some possible directions for future work, where one could focus on improving \ours's predictions under such strict thresholds. 
\smalljump 

\noindent\textbf{Pose Solvers.} Arnold \etal \cite{arnold2022map} propose different strategies for recovering the metric scale from the keypoint correspondences. In the first strategy, authors first compute the essential matrix and then rely on the depth estimation to obtain the scaled translation vector (Ess. Scale)~\cite{nister2004efficient, fischler1981random}. Their second strategy consisted of using the depth maps to lift the 2D keypoints to 3D, and then applying the Perspective-n-Point (PnP) algorithm \cite{gao2003complete}. We refer to \cite{arnold2022map} for more details. We compare such strategies in Table \ref{tab:sup:pose_ablation}. Moreover, we also show \ours's results with the two different proposed solvers. We demonstrate that obtaining poses with the same probabilistic approach we use during training yields the best results, proving the effectiveness of both, the end-to-end strategy and our probabilistic formulation of the metric relative pose estimation. 
\smalljump 

\begin{table}[t]
\footnotesize
\begin{center}
\begin{tabular}{c l c c l c c}
\multicolumn{7}{c}{\textbf{Map-free Dataset}}\\ 
\cline{1-7}
\noalign{\smallskip}
\cline{1-7}
\noalign{\smallskip}
\multicolumn{2}{c}{} & \multicolumn{2}{c}{VCRE (90px)} &\multicolumn{1}{c}{} & \multicolumn{2}{c}{Pose (25cm, 5\textdegree)}\\ 
\cline{3-4} \cline{6-7} \noalign{\smallskip}
\multicolumn{2}{c}{} & \multicolumn{1}{c}{AUC} & \multicolumn{1}{c}{Prec. (\%)} & \multicolumn{1}{c}{} & \multicolumn{1}{c}{AUC} & \multicolumn{1}{c}{Prec. (\%)}\\ 
\cline{1-7}
\noalign{\smallskip}
\textbf{Pose Solvers} &&&& \\
\cline{1-7}
\noalign{\smallskip}
\textbf{SuperGlue}~\cite{detone2018superpoint, sarlin2020superglue} &&&& \\
\cline{1-1} \noalign{\smallskip}
Ess. Scale && \textbf{0.60} & \textbf{36.1} && \textbf{0.35} & \textbf{16.8} \\
PnP && \textbf{0.60} & 36.0 && 0.25 & 10.7 \\
\cline{1-7}
\noalign{\smallskip}
\textbf{LoFTR}~\cite{sun2021loftr} &&&& \\
\cline{1-1} \noalign{\smallskip}
Ess. Scale && 0.61 & \textbf{34.7} && \textbf{0.35} & \textbf{15.4} \\
PnP && \textbf{0.62} & 33.4 && 0.27 & 9.8 \\
\cline{1-7}
\noalign{\smallskip}
\textbf{\ours w/ Overlap} &&&& \\
\cline{1-1} \noalign{\smallskip}
Ess. Scale && 0.66 & 39.3 && 0.22 & 8.4 \\
PnP && 0.70 & 42.1 && \textbf{0.36} & \textbf{14.6} \\
\textbf{Our Solver} && \textbf{0.75} & \textbf{49.2} && 0.33 & 13.3 \\
\cline{1-7}
\noalign{\smallskip}
\textbf{\ours} &&&& \\
\cline{1-1} \noalign{\smallskip}
Ess. Scale && 0.65 & 37.1 && 0.20 & 6.9 \\
PnP && 0.70 & 42.5 && \textbf{0.33} & \textbf{12.8} \\
\textbf{Our Solver} && \textbf{0.74} & \textbf{49.2} && 0.28 & 12.0 \\
\end{tabular}
\end{center}
\vspace{-2em}
\normalsize
\caption{\textbf{Pose solver ablation on Map-free}.
Results show that state-of-the-art matchers work better when estimating the essential matrix from 2D-2D correspondences, and then recovering the metric scale from the depth predictor. \ours, meanwhile, obtains the top VCRE results when recovering the pose with the probabilistic solver used during training. 
}
\label{tab:sup:pose_ablation}
\end{table}

\noindent\textbf{Cross-dataset evaluation.} We also test the generalization capability of \ours when trained and tested in different scenarios. We use \ours trained in ScanNet dataset and evaluate it in the Map-free evaluation. Even though this experiment involves a significant distribution gap (indoor vs.~outdoor), \ours achieves an AUC (VCRE) score of 0.55, still outperforming DISK \cite{tyszkiewicz2020disk}, SiLK \cite{gleize2023silk}, and DeDoDe \cite{edstedt2023dedode}, which all were trained on outdoor datasets (see Table \ref{tab:sup:mapfree_additional} for all AUC (VCRE) results).  
\smalljump

\begingroup

\setlength{\tabcolsep}{4pt} % Default value: 6pt

\begin{table}[t]
\footnotesize
\begin{center}
\begin{tabular}{cccclccc}
\multicolumn{1}{c}{} & \multicolumn{3}{c}{\textbf{DIML Outdoor} \cite{kim2018deep} } & \multicolumn{1}{}{} & \multicolumn{3}{c}{\textbf{DIODE Outdoor} \cite{vasiljevic2019diode}}\\
\hline \noalign{\smallskip}
% \cline{2-4} \cline{6-8} \noalign{\smallskip}
\multicolumn{1}{c}{} & \multicolumn{1}{c}{$\delta_1$ $\uparrow$} & \multicolumn{1}{c}{REL $\downarrow$} & \multicolumn{1}{c}{RMSE $\downarrow$}& \multicolumn{1}{}{} & \multicolumn{1}{c}{$\delta_1$ $\uparrow$} & \multicolumn{1}{c}{REL $\downarrow$} & \multicolumn{1}{c}{RMSE $\downarrow$}\\
\hline \noalign{\smallskip}
% KBR & 0.57 ($\delta_{0.25}$) & 33.49 & x & x \\
% ZoeD-K & 0.003 ($\delta_{1}$) & 1.921 & 6.978 & - \\
ZoeDepth~\cite{bhat2023zoedepth} & 0.29 & 0.64 & 3.61 && \textbf{0.21} & 0.76 & \textbf{7.57} \\
\textbf{\ours} & 0.65 & 0.20 & 4.30 && 0.04 & 0.67 & 15.18 \\
\textbf{\ours-Sc} & \textbf{0.70} & \textbf{0.17} & \textbf{2.39} && 0.04 & \textbf{0.66} & 13.76 \\
\end{tabular}
\end{center}
\vspace{-2.0em}
\normalsize
\caption{\textbf{Outdoor monocular depth evaluation.} We report the zero-shot generalization on outdoor datasets and see that \ours provides competitive results even though it was not designed for this task.}
\label{sup:tab:depth_outdoor}
\vspace{-0.7em}
\end{table}
\endgroup

\noindent\textbf{Monocular depth estimation.} In Table \ref{sup:tab:depth_outdoor}, we further evaluate the generalization capabilities of our network also in the zero-shot monocular depth estimation task. We compute its accuracy in the DIML Outdoor~\cite{kim2018deep} and the DIODE Outdoor~\cite{vasiljevic2019diode} datasets. As a reference, we also provide ZoeDepth (NK) metrics. Similar to Section \ref{sup:exp:scannet} (Table \ref{tab:sup:monodepth_scannet}), we also show results for \ours-Sc, where we evaluate depth prediction on the positions where \ours is most confident (50\% top scoring positions). \ours has been trained with pedestrian smartphone images, and still, it can generalize and produce valid depth maps in datasets with different statistics, or visual conditions. 
\smalljump

\begin{table}[t]
\footnotesize
\begin{center}
\begin{tabular}{c l c l c}
\multicolumn{5}{c}{\textbf{Map-free Dataset}}\\ 
\cline{1-5}
\noalign{\smallskip}
\cline{1-5}
\noalign{\smallskip}
\multicolumn{2}{c}{} & \multicolumn{1}{c}{VCRE} & \multicolumn{1}{c}{} & \multicolumn{1}{c}{Median Errors}\\ 
\cline{3-3} \cline{5-5} \noalign{\smallskip}
\multicolumn{2}{c}{} & \multicolumn{1}{c}{AUC / Prec. (\%)} & \multicolumn{1}{c}{} & \multicolumn{1}{c}{Rep. / Trans / Rot}\\ 
\cline{1-5}
\noalign{\smallskip}
\textbf{Hard pairs} && \\
\cline{1-1} \noalign{\smallskip}
SuperGlue \cite{detone2018superpoint, sarlin2020superglue} && 0.07 / 3.5 && 271.8 / 4.5 / 81.6 \\
LoFTR \cite{sun2021loftr} && 0.06 / 2.6 && 255.7 / 4.8 / 88.5 \\
RoMa \cite{edstedt2023roma} && \underline{0.08} / \underline{7.3} && 241.7 / \textbf{3.1} / \textbf{58.3} \\
\textbf{MicKey} (ours) && \textbf{0.15} / \textbf{11.1} && \textbf{233.7} / \underline{3.7} / \underline{75.2} \\
\end{tabular}
\end{center}
\vspace{-2em}
\normalsize
\caption{\textbf{Hard examples on the Map-free dataset.}
We evaluate image pairs from the validation set that are taken under large viewpoint changes. We define such examples as image pairs that are at least 3m apart and have a 45\textdegree{} change in the camera direction.
We report the VCRE metrics and the median errors of the estimated poses.}
\label{tab:sup:diff_pairs}
\end{table}
\begin{figure*}[t]
  \centering
   \includegraphics[width=1.0\linewidth]{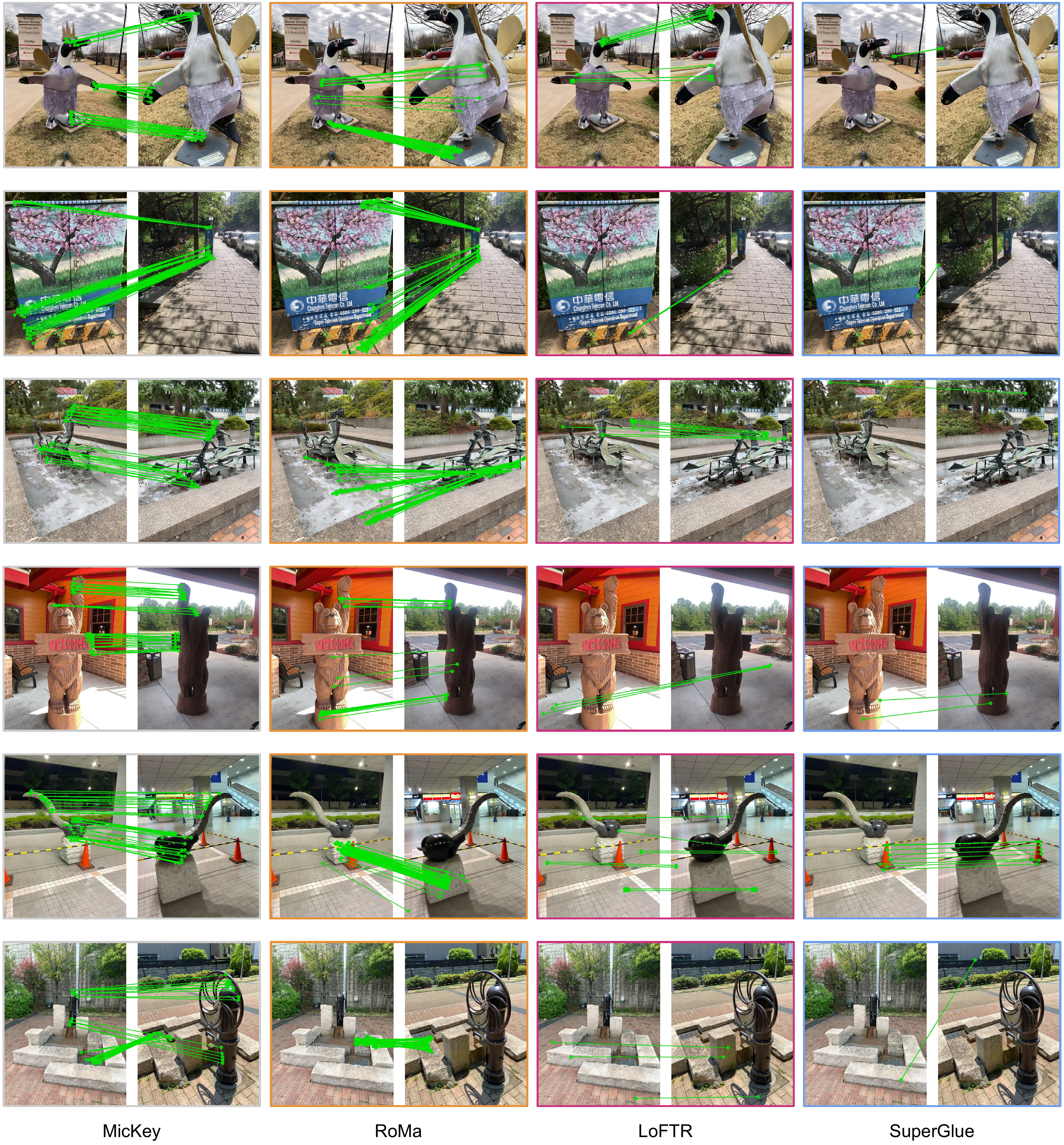}
   \vspace{-1em}
   \caption{\textbf{Inlier correspondences on Map-free dataset.} 
     We show the inlier correspondences returned by different feature extractors and their pose solvers. \ours outperforms the other matchers when there are strong viewpoint changes between the two input images. 
     % We observe that \ours is able to hallucinate the keypoint correspondences. 
     \ours embeds in a single neural network a feature representation of the keypoints, as well as their 3D geometry, allowing it to match keypoints where little overlap is observed. 
     We see that SuperGlue~\cite{detone2018superpoint, sarlin2020superglue}, a sparse feature method like \ours, struggles to compute good correspondences  when images present this kind of extreme viewpoint differences. 
   }
      \vspace{1em}
   \label{fig:sub:mapfree_visual}
\end{figure*}
\noindent\textbf{Inlier correspondences.}
In this last section, we show different visual examples in Figure \ref{fig:sub:mapfree_visual}. We plot the inlier correspondences that every method returns after computing the relative pose estimation. We observe that \ours finds correct correspondences even though images were taken from extremely different viewpoints. Moreover, we see that \ours detects and tries to match the object of interest within the image instead of relying on local patterns that might not appear in the two images. 
For instance, in images from row 1 or row 4, the object of interest is shown in the images from opposite views, \ie, images were taken with almost a  180\textdegree{} difference. Even so, \ours is able to match the correct side of the object to its corresponding part in the other image. 
Thus, we observe that the network is able to reason about the shape of the object and establish correspondences beyond local patterns. RoMa~\cite{edstedt2023roma} is also able to find good matches, but it fails when the images do not have direct visual overlap. 
Contrary to LoFTR~\cite{sun2021loftr} or RoMA~\cite{edstedt2023roma}, \ours and SuperGlue~\cite{sarlin2020superglue} are sparse feature methods, and then they only have access to a single image when computing their keypoints and descriptors. Contrary to \ours, we note that state-of-the-art sparse feature methods (\eg, SuperPoint~\cite{detone2018superpoint}-SuperGlue~\cite{sarlin2020superglue}) do not find any good correspondences under such extreme cases.

We report the VCRE metrics and median errors on image pairs that have large and challenging viewpoint differences in Table \ref{tab:sup:diff_pairs}. We use the validation scenes, where ground truth data is provided and hence, we can define the difficulty of an image pair. We rely on the pose difference between the reference and the query frame instead of the overlap score, such that unsolvable pairs are also evaluated. We define a hard example as a pair that is taken at least 3 meters apart and with their camera directions being at least rotated by 45\textdegree{}. We see that on those challenging examples, \ours obtains the highest number of relative poses under the VCRE threshold (90 pixels).

\end{appendices}

{
    \small
    \bibliographystyle{ieeenat_fullname}
    \bibliography{main}
}

% WARNING: do not forget to delete the supplementary pages from your submission 
% \input{sec/X_suppl}

\end{document}